%% file: iclr2026_conference.tex
\documentclass{article} % For LaTeX2e

\usepackage{iclr2026_conference,times}
\usepackage{hyperref}
\usepackage{url}
\usepackage{graphicx}
\usepackage{float}
\usepackage{etoolbox}
\setkeys{Gin}{width=\textwidth}
\AtBeginEnvironment{figure}{\centering}
\usepackage{todo}
\usepackage[utf8]{inputenc}
\usepackage[T1]{fontenc}
\usepackage{booktabs}
\usepackage{longtable}
\usepackage{array}
\usepackage{hyperref}
\usepackage{geometry}
\usepackage{xcolor}
\usepackage{multirow}
\usepackage{caption}
\usepackage{tabularx}
\usepackage{array}
\usepackage{pifont}
\usepackage[table]{xcolor}
\usepackage{booktabs}
\usepackage{tikz}

\usepackage{tabularx}
\usepackage{tcolorbox}
\tcbuselibrary{breakable,skins}

\newtcolorbox{samplebox}[1]{%
  enhanced,
  breakable,
  colback=white,
  colframe=black!25,
  boxrule=0.4pt,
  arc=1.2mm,
  left=6pt,right=6pt,top=4pt,bottom=4pt,
  fonttitle=\bfseries\footnotesize,
  title={#1}%
}

\usetikzlibrary{positioning,shapes.geometric,arrows.meta}

\definecolor{good}{RGB}{183,224,117}   % green
\definecolor{mid}{RGB}{255,255,255}    % white
\definecolor{bad}{RGB}{248,180,180}    % red

\newcommand{\SCORE}[1]{\ensuremath{S_{\textrm{#1}}}}

\newcommand{\QAtext}[2]{\ensuremath{QA_{\text{#1}}^{\text{#2}}}}
\newcommand{\QA}[2]{\QAtext{#1}{#2}}

\newcommand{\QAcode}[2]{\ensuremath{QA_{#1}^{#2}}}
\newcommand{\QAcodeText}[2]{\ensuremath{QA_{#1}^{\text{#2}}}}

\newcommand{\QApipe}[2]{\textbf{$\QAcode{#1}{#2}$}}

\newcommand{\sampleblock}[1]{%
  \par\noindent
  \begin{minipage}{\linewidth}\small\ttfamily
    #1
  \end{minipage}\par
}

\input{math_commands.tex}

\title{DISCO: Document Intelligence Suite for COmparative Evaluation}

\author{
  \begin{tabular}[t]{@{}l@{\hspace{1.2cm}}l@{}}
    \begin{tabular}[t]{@{}l@{}}
      \textbf{Kenza Benkirane} \\
      {\normalfont Parexel AI Labs} \\
      {\normalfont London, United Kingdom} \\
      {\normalfont \texttt{kenza.benkirane@parexel.com}}
    \end{tabular}
    &
    \begin{tabular}[t]{@{}l@{}}
      \textbf{Dan Goldwater} \\
      {\normalfont Parexel AI Labs} \\
      {\normalfont London, United Kingdom} \\
      {\normalfont \texttt{dan.goldwater@parexel.com}}
    \end{tabular}
    \\[4em]
    \begin{tabular}[t]{@{}l@{}}
      \textbf{Martin Asenov} \\
      {\normalfont Parexel AI Labs} \\
      {\normalfont London, United Kingdom} \\
      {\normalfont \texttt{martin.asenov@parexel.com}}
    \end{tabular}
    &
    \begin{tabular}[t]{@{}l@{}}
      \textbf{Aneiss Ghodsi} \\
      {\normalfont Parexel AI Labs} \\
      {\normalfont San Francisco, United-States} \\
      {\normalfont \texttt{aneiss.ghodsi@parexel.com}}
    \end{tabular}
  \end{tabular}
}

\iclrfinalcopy % Uncomment for camera-ready version, but NOT for submission.
\begin{document}

\maketitle

\begin{abstract}
  Document intelligence requires accurate text extraction and reliable reasoning over document content. We introduce \textbf{DISCO}, a \emph{Document Intelligence Suite for COmparative Evaluation}, that evaluates optical character recognition (OCR) pipelines and vision-language models (VLMs) separately on parsing and question answering across diverse document types, including handwritten text, multilingual scripts, medical forms, infographics, and multi-page documents. Our evaluation shows that performance varies substantially across tasks and document characteristics, underscoring the need for complexity-aware approach selection. OCR pipelines are generally more reliable for handwriting and for long or multi-page documents, where explicit text grounding supports text-heavy reasoning, while VLMs perform better on multilingual text and visually rich layouts. Task-aware prompting yields mixed effects, improving performance on some document types while degrading it on others. These findings provide empirical guidance for selecting document processing strategies based on document structure and reasoning demands.
  \footnote{
    % GitHub: \url{https://github.com/kenza-ily/disco} \\
  Hugging-Face: \url{https://huggingface.co/collections/kenza-ily/disco}}

\end{abstract}

\section{Introduction and Related Work}

Documents remain a primary source of information across industries, yet extracting and reasoning over their content poses persistent challenges. Traditional approaches rely on OCR to convert images to text, followed by language models for downstream tasks. Recent VLMs offer an alternative: processing document images directly without explicit text extraction. This raises a practical question: when should practitioners use OCR pipelines versus end-to-end VLMs? Current benchmarks report only final task accuracy, making it difficult to diagnose whether failures stem from text extraction or reasoning.

\noindent\textbf{Existing benchmarks and their gaps.} Document understanding evaluation has evolved from isolated OCR benchmarks to integrated question answering (QA) datasets. DocVQA~\citep{mathew2021docvqa} requires reading and reasoning over forms and letters; InfographicVQA~\citep{mathew2022infographicvqa} targets visual reports; DUDE~\citep{vanlandeghem2023dude} introduces multi-page, multi-domain documents. ChartQAPro~\citep{masry-etal-2025-chartqapro} revealed that performance on narrow benchmarks overestimates generalisation: models achieving $\sim$90\% on ChartQA dropped to $\sim$56\% on more diverse charts. Despite this progress, benchmarks typically report end-to-end accuracy without isolating error sources. When a system fails, we cannot determine whether OCR missed the text, the layout was misinterpreted, or the reasoning was flawed.

\noindent\textbf{OCR pipelines versus VLMs.} OCR-first pipelines benefit from decades of optimisation for text recognition and can handle long documents efficiently, but may lose spatial relationships when converting content to plain text. VLMs process images holistically, preserving layout cues, but require high resolution to read fine print and may hallucinate when text is unclear. Recent hybrid approaches, such as DocVLM~\citep{NacsonAGBGKML25}, inject OCR-detected text as additional tokens into frozen VLMs and achieve strong results on DUDE. This suggests that the two paradigms may be complementary rather than competing; however, systematic comparison across document types remains lacking.

\noindent\textbf{Our contribution.} We introduce \textsc{DISCO}, a diagnostic evaluation framework for document intelligence that explicitly separates text parsing from downstream question answering. Rather than treating document understanding as a single end-to-end prediction problem, DISCO evaluates intermediate representations and final answers under controlled pipeline variants, enabling attribution of errors to perception, representation, or reasoning stages.

Using this stage-wise protocol, we report three empirical observations. First, model behaviour depends strongly on document structure: OCR-based pipelines are generally more reliable for long and multi-page documents, while VLM-based approaches tend to be stronger on multilingual text and visually structured content such as infographics and forms. Second, task-aware prompting has heterogeneous effects across datasets and model families, helping on some document types but offering limited gains on challenging domains such as medical prescriptions. Third, direct visual question answering can be advantageous on single-page documents, suggesting that intermediate text extraction may introduce information loss when spatial layout is central to the task.
% Critically, we find that parsing quality establishes a performance ceiling: on multi-page documents, a 10\% improvement in OCR coverage translates to approximately +3 percentage points in QA accuracy (Section~\ref{sec:ocr-ceiling}), whilst on visually grounded documents, direct VQA bypasses this bottleneck entirely.

Overall, DISCO reframes document intelligence evaluation as turns end-to-end accuracy into actionable diagnosis, revealing not just which system wins, but why it wins. By analysing intermediate representations in addition to final predictions, our framework highlights the limits of assessing multimodal systems solely through end-to-end accuracy and motivates evaluation beyond next-token prediction, particularly for documents where perception, layout, and reasoning interact.

\section{Datasets and benchmark suite}
% We curate compact but diverse subsets from established benchmarks covering handwritten text (IAM, \citet{marti2002iam}), multilingual scene text (ICDAR, \citet{karatzas2015icdar}), medical forms (RxPad, \citet{pattin2026rxpad}; scanned documents (DocVQA, \citet{mathew2021docvqa}), infographics (InfographicVQA, \citet{mathew2022infographicvqa}), and heterogeneous multi-page documents (DUDE, \citet{vanlandeghem2023dude}). Each subset contains fewer than 500 samples, allowing consistent evaluation across tasks while preserving document diversity.

\begin{table*}[t]
  \centering
  \tiny
  \setlength{\tabcolsep}{4pt}
  \begin{tabularx}{0.9\textwidth}{
      >{\centering\arraybackslash}p{4cm}
      >{\centering\arraybackslash}p{4.5cm}
      >{\raggedright\arraybackslash}X
    }
    \toprule
    \textbf{Task} & \textbf{Dataset} & \textbf{Description} \\
    \midrule
    Parsing & \textbf{IAM}$_{\text{DISCO}}$ from~\cite{marti2002iam} & Handwritten text \\
    Parsing & \textbf{ICDAR 2015}$_{\text{DISCO}}$ from~\cite{karatzas2015icdar} & Multilingual scene text \\
    Parsing & \textbf{PubLayNet}$_{\text{DISCO}}$ from~\cite{zhong2019publaynet} & Scientific document pages (layout) \\
    Parsing & \textbf{RxPad}$_{\text{DISCO}}$ from~\cite{pattin2026rxpad} & Medical forms (French) \\
    QA & \textbf{DocVQA}$_{\text{DISCO}}$ from~\cite{mathew2021docvqa} & Scanned forms and letters \\
    QA & \textbf{InfographicVQA}$_{\text{DISCO}}$ from~\cite{mathew2022infographicvqa} & Infographics \\
    QA & \textbf{DUDE}$_{\text{DISCO}}$ from~\cite{vanlandeghem2023dude} & Heterogeneous multi-page documents \\
    QA & \textbf{ChartQAPro}$_{\text{DISCO}}$ from~\cite{masry2025chartqapro} & Charts \\
    QA & \textbf{VisR-Bench}$_{\text{DISCO}}$ from~\cite{chen2025visr-bench} & Long multi-page documents \\
    \bottomrule
  \end{tabularx}
  \caption{Benchmark suite composition: datasets and document types evaluated in DISCO. \textit{Note:} All DISCO dataset subsets and accompanying evaluation artifacts will be released upon paper acceptance and with the final version of the paper. VisR-Bench and PubLayNet are included in the suite, but we do not report experimental results on them in this paper.}
  \label{tab:datasets}
\end{table*}

We construct a benchmark suite designed to evaluate document intelligence systems across two core tasks: text parsing and question answering. Rather than relying on a single dataset, we combine multiple established benchmarks to ensure coverage of diverse document types, visual characteristics, languages, and reasoning requirements. This choice reflects the heterogeneous nature of real-world documents and allows us to assess models under a wide range of conditions. By evaluating across these datasets, we aim to capture both low-level perception challenges and higher-level reasoning and grounding behaviour. The datasets are summarised in Table~\ref{tab:datasets}.

To ensure feasibility and reproducibility, we restrict each dataset to fewer than 500 samples. For large-scale benchmarks, we construct dedicated \textit{DISCO} versions by sampling small but representative subsets. Sampling is performed to preserve key properties of the original datasets; including document structure, question and answer types, language distribution, and difficulty. When metadata is available, stratified sampling is used to avoid bias towards simpler instances. Datasets with fewer than 500 samples are used in full.

This design enables consistent evaluation across tasks while keeping computational cost manageable. It also allows us to apply a unified experimental protocol and identical model configurations across datasets, facilitating fair comparisons between approaches.

A summary of all datasets, their associated tasks, and their main characteristics is provided in Table~\ref{tab:datasets}. Detailed dataset descriptions, sampling procedures, and statistics for the \textit{DISCO} versions are reported in Appendix~\ref{appendix:representative_suite}.

\section{Methodology and experimental design}

We evaluate document intelligence by separating parsing from question answering, testing both OCR systems and VLMs with prompt variations\footnote{Prompts are available in Section~\ref{sec:prompts}.} to assess capabilities at each stage.
Throughout, we write metric scores using the notation $\SCORE{M}$, where $M$ is the metric name (e.g., $\SCORE{CS}$ for cosine similarity). In each condition, we compare OCR systems (\texttt{azure-ai-documentintelligence}, \texttt{mistral-ocr-2505}) against VLMs (\texttt{gpt-5-mini}, \texttt{gpt-5-nano}, \texttt{claude-3-5-sonnet}). All experiments use deterministic decoding and a fixed image resolution for fair comparison.

\textbf{Text parsing} extracts textual content from document images, including printed text, handwriting, multilingual scripts, and structured layouts. We evaluate on three datasets: IAM$_{\text{DISCO}}$ (handwriting; \citet{marti2002iam}), ICDAR$_{\text{DISCO}}$ (multilingual scene text; \citet{karatzas2015icdar}), RxPad (\citet{pattin2026rxpad}). Our experiments are as follows: \textbf{$P_{\text{OCR}}$} (OCR-only parsing), \textbf{$P_{\text{VLM-base}}$} (VLM base prompt parsing), and \textbf{$P_{\text{VLM-task}}$} (parsing with task-aware prompt). We report $\SCORE{CS}$ (cosine similarity between embeddings of extracted and ground-truth text; higher is better), $\SCORE{CER}$ (character error rate: normalised character-level edit distance; lower is better), and $\SCORE{WER}$ (word error rate: normalised word-level edit distance; lower is better).

\textbf{Question answering} evaluates answer generation from document content. We test on DocVQA$_{\text{DISCO}}$ (forms; \citet{mathew2021docvqa}), InfographicVQA$_{\text{DISCO}}$ (infographics; \citet{mathew2022infographicvqa}) and DUDE$_{\text{DISCO}}$ (multi-page documents; \citet{vanlandeghem2023dude}). Our experiments are as follows: \textbf{$QA_{\text{OCR}}$} (OCR parsing $\to$ LLM QA), \textbf{$QA_{\text{VLM-2stage}}$} (VLM parsing $\to$ LLM QA; two-stage), and \textbf{$QA_{\text{VLM-direct}}$} (direct VLM QA). Throughout, we use the convention $QA_{\text{OCR}/\text{VLM-2stage}/\text{VLM-direct}}^{\text{generic}/\text{cot}/\text{task-aware}}$, where the subscript denotes the pipeline and the superscript denotes the prompt, e.g., \textbf{$QA_{\text{OCR}}^{\text{generic}}$}, \textbf{$QA_{\text{OCR}}^{\text{cot}}$}, and \textbf{$QA_{\text{OCR}}^{\text{task-aware}}$}. For the parsing in the QA tasks, we also use \texttt{mistral-ocr-2512}. Each pipeline is tested with simple, detailed, and context-aware prompts. We report $\SCORE{GT-in-Pred}$ (ground-truth-in-prediction: substring match indicator; higher is better), $\SCORE{ANLS}$ (average normalised Levenshtein similarity; higher is better), and $\SCORE{EM}$ (exact match rate; higher is better).

\section{Results and discussion}

\begin{table}[h]
  \centering
  \small
  \setlength{\tabcolsep}{6pt}
  \begin{tabular}{llccc}
    \toprule
    \textbf{Task} & \textbf{Dataset} &
    \textbf{$P_{\text{OCR}}$} &
    \textbf{$P_{\text{VLM-base}}$} &
    \textbf{$P_{\text{VLM-task}}$} \\
    \midrule
    \multirow{3}{*}{Parsing - $\SCORE{CER}$}
    & IAM$_{\text{DISCO}}$    & 0.087 & 0.171 & \textbf{0.080} \\
    & ICDAR$_{\text{DISCO}}$  & 0.553 & 0.213 & \textbf{0.073} \\
    & RxPad  & \textbf{0.654} & 0.660 & 0.659 \\
    \midrule
    \textbf{Task} & \textbf{Dataset} &
    \textbf{$QA_{\text{OCR}}$} &
    \textbf{$QA_{\text{VLM-2stage}}$} &
    \textbf{$QA_{\text{VLM-direct}}$} \\
    \midrule
    \multirow{3}{*}{QA - $\SCORE{GT-in-Pred}$}
    & DocVQA$_{\text{DISCO}}$         & 0.876 & 0.896 & \textbf{0.908} \\
    & InfographicVQA$_{\text{DISCO}}$ & 0.754 & 0.711 & \textbf{0.785} \\
    & DUDE$_{\text{DISCO}}$           & \textbf{0.562} & 0.555 & 0.498 \\
    \bottomrule
  \end{tabular}
  \caption{DISCO benchmark results - Parsing $\SCORE{CER}$ and QA $\SCORE{GT\text{-}in\text{-}Pred}$ across OCR a VLM pipeline variants.}
  \label{tab:results-combined}
\end{table}

\subsection{Parsing}

\textbf{OCR performs best on handwriting unless VLMs use task-aware prompting.}
On IAM$_{\text{DISCO}}$, OCR achieved low $\SCORE{CER}$ (0.087--0.089), while generic VLM prompts performed substantially worse. Task-aware prompting closed this gap and slightly outperformed OCR ($\SCORE{CER}$ 0.080). At the word level, VLMs showed lower $\SCORE{WER}$ than OCR, suggesting fewer word-level errors despite weaker character accuracy with generic prompts.

\textbf{VLMs outperform OCR on multilingual scene text.}
For ICDAR$_{\text{DISCO}}$, VLMs clearly surpassed OCR. Generic prompting reduced $\SCORE{CER}$ from 5.53\% to 2.13\%, while task-aware prompting further improved performance to 0.73\%. This pattern is consistent with strong multilingual pre-training in VLMs.

\textbf{Medical prescriptions remain challenging.}
On RxPad, all methods showed similarly high error rates. OCR and VLMs achieved nearly identical $\SCORE{CER}$ and $\SCORE{WER}$, with only marginal gains from task-aware prompts. This indicates persistent challenges related to handwriting variation, domain terminology, and layout.

\subsection{Question answering}

\textbf{Direct VQA performs best on single-page documents.}
On DocVQA$_{\text{DISCO}}$, direct VQA ($\QAtext{VLM-direct}{task-aware}$) achieved the highest score (0.908), outperforming both OCR-based ($\QAtext{OCR}{task-aware}$) and text-based VLM pipelines ($\QAtext{VLM-2stage}{task-aware}$). Avoiding intermediate text extraction appears to reduce error propagation.

\textbf{Visual structure matters for infographics.}
A similar trend appears on InfographicVQA$_{\text{DISCO}}$, where $\QAtext{VLM-direct}{task-aware}$ outperformed $\QAtext{OCR}{task-aware}$ and $\QAtext{VLM-2stage}{task-aware}$. The weaker performance of $\QAtext{VLM-2stage}{task-aware}$ suggests that linearised text representations lose spatial and visual cues that are important for infographic understanding. In cases where models achieve high $\SCORE{GT-in-Pred}$ but low $\SCORE{ANLS}$ or $\SCORE{EM}$, this reflects answer localisation with non-conforming output format, rather than improved scores due to verbosity.

\textbf{OCR pipelines remain competitive for long documents but model selection remains important.}
On DUDE$_{\text{DISCO}}$, the OCR-based pipeline achieved the best performance. Direct VQA underperformed, highlighting current limitations of VLMs on longer contexts even with controlled evidence-page access. However, on DocVQA$_{\text{DISCO}}$, \texttt{azure-ai-documentintelligence} outperformed \texttt{mistral-ocr-2505} by 3.3 percentage points (0.876 vs 0.843), indicating that single-page form documents benefit from Azure's stronger layout analysis.

\subsection{Conclusion and discussion}

The results support a dual strategy: OCR-based pipelines are more reliable for complex text, long documents, and text-heavy reasoning, where structured textual representations and controlled retrieval are essential, while VLM-based end-to-end approaches are better suited to visually grounded documents, such as infographics and natural scene text, where spatial layout and visual cues play a central role. When performance differences are small, VLM pipelines offer an additional practical advantage by avoiding OCR-induced error propagation and simplifying system design. These results suggest that document structure and layout complexity, rather than model family alone, should guide the choice between OCR-based and end-to-end multimodal pipelines.

\clearpage
\bibliography{iclr2026_conference}
\bibliographystyle{iclr2026_conference}

\clearpage

\section*{APPENDIX}

\appendix

\section{Limitations and future work}
\label{ap:limitations}

\textbf{Retrieval and long-context reasoning} Our evaluation focuses on contexts where full documents can be processed directly. We do not evaluate retrieval mechanisms, which are essential for practical long-document systems. In real-world applications, documents often span dozens or hundreds of pages (e.g., clinical study reports, financial statements, insurance claims), requiring retrieval to locate relevant passages before answering questions. Future work should assess retrieval-augmented pipelines, comparing vector-based text retrieval against vision-based page selection to understand where each approach succeeds or fails.

\textbf{Multilingual and non-Latin script coverage} While ICDAR$_{\text{DISCO}}$ revealed the need for better multilingual support, our suite primarily covers English and French text in Latin scripts.  Future benchmarks should systematically evaluate non-Latin scripts, mixed-script documents, and culturally specific layouts to guide deployment in global healthcare and regulatory contexts.

\textbf{Metric limitations and answer format variability} We deliberately used deterministic metrics ($\SCORE{GT-in-Pred}$, $\SCORE{ANLS}$, $\SCORE{EM}$) rather than LLM-as-judge evaluation to ensure reproducible, non-stochastic assessment across all experiments. However, this approach requires considering multiple metrics together to see the full picture. Models frequently located correct information (high $\SCORE{GT-in-Pred}$) but failed to format answers appropriately (low $\SCORE{ANLS}$ or $\SCORE{EM}$), revealing a gap between answer localisation and output formatting. Future work could explore hybrid approaches that maintain reproducibility whilst better capturing semantic equivalence—for example, few-shot prompting with answer format examples or structured output validation using Pydantic schemas to constrain response formats without introducing evaluation non-determinism.

\textbf{Verbosity and information completeness} VLMs often produced longer outputs than ground-truth references, particularly during parsing. Qualitative inspection suggested these were not always hallucinations but rather more verbose extractions including contextual information. Topic-specific evaluation frameworks could assess what types of information are consistently missed or fabricated—for example, whether dosage errors in prescriptions are more common than patient name errors, or whether financial figures in regulatory documents are more prone to hallucination than procedural descriptions.

% \subsubsection*{Author Contributions}
% If you'd like to, you may include  a section for author contributions as is done
% in many journals. This is optional and at the discretion of the authors.

% \subsubsection*{Acknowledgments}

% \pagebreak
\clearpage
\section{Full experimental results}
\label{appendix:detailed_results}

%==============================================================================
\subsection{Parsing task: in-depth analysis}
\label{appendix:parsing_analysis}

\subsubsection{Handwriting recognition (IAM$_{\text{DISCO}}$)}

IAM$_{\text{DISCO}}$ contains 500 handwritten text samples with varying writing styles. VLMs with task-aware prompting can match or outperform dedicated OCR systems on character-level accuracy: \texttt{gpt-5-mini} achieves $\SCORE{CER} = 0.080$, compared to $\SCORE{CER} = 0.087$ for \texttt{mistral-ocr-2505}. The word-level gap is more pronounced---\texttt{gpt-5-mini} reaches $\SCORE{WER} = 0.110$ versus $0.305$ for OCR, nearly $3\times$ better. However, OCR systems maintain higher semantic fidelity, with \texttt{azure-ai-documentintelligence} achieving $\SCORE{CS} = 0.946$ versus $0.914$ for the best VLM. This suggests OCR errors are more localised (character substitutions preserving meaning), while VLM errors may involve rephrasing that shifts semantic representation.

Task-aware prompting yields inconsistent effects across model families. \texttt{gpt-5-mini} improves from $\SCORE{CS} = 0.827$ (generic) to $0.914$ (task-aware), a $+10.5\%$ gain, with $\SCORE{CER}$ dropping from $0.175$ to $0.080$. \texttt{claude-3-5-sonnet} exhibits the opposite pattern: performance degrades from $\SCORE{CS} = 0.905$ to $0.845$, and $\SCORE{CER}$ increases from $0.163$ to $0.201$. This divergence indicates that handwriting-specific instructions may conflict with certain models' default transcription behaviour, and prompting strategies cannot be assumed to transfer across model families.

\begin{figure}[h]
  \centering
  \includegraphics[width=\textwidth]{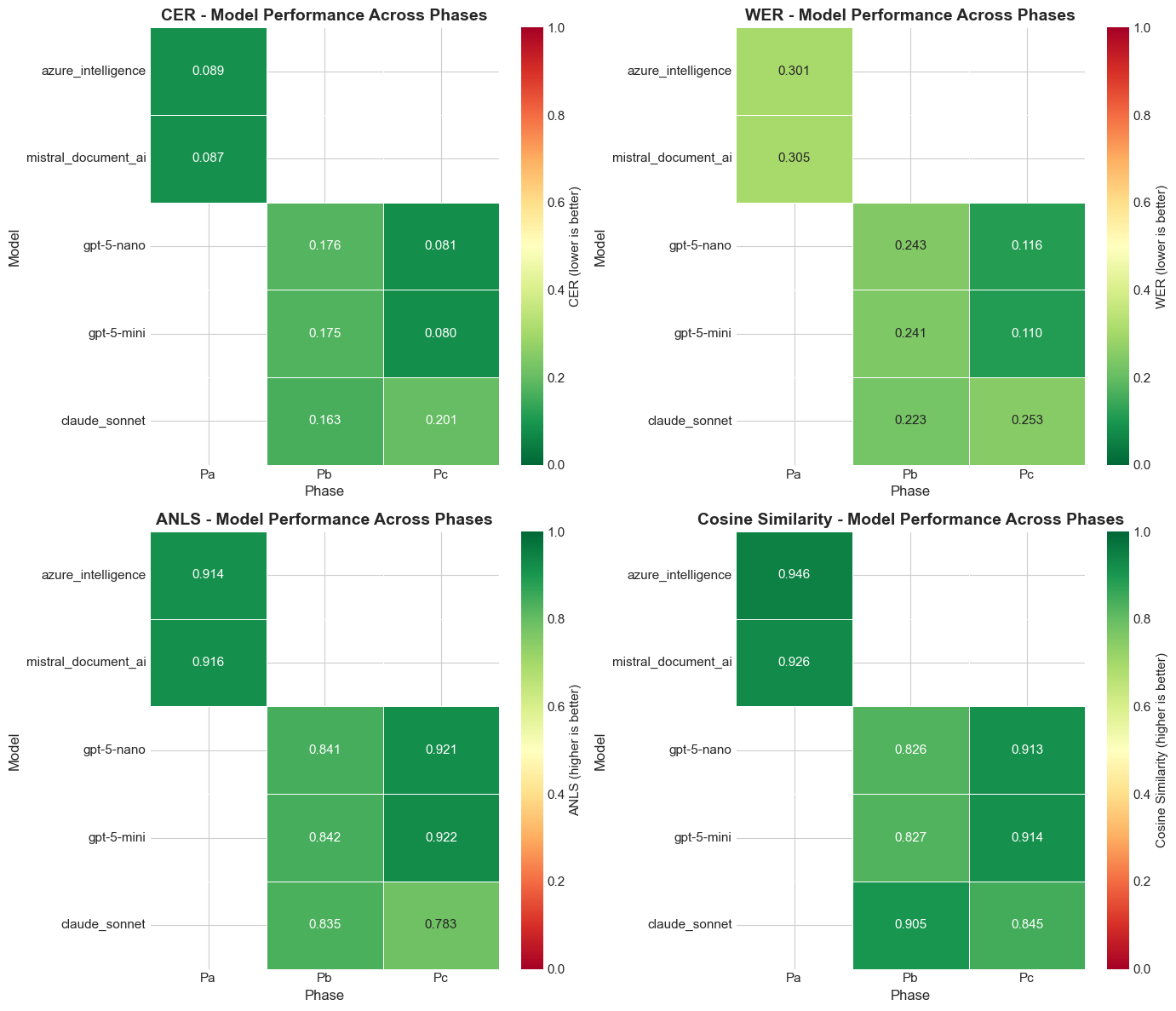}
  \caption{Model performance across phases on IAM$_{\text{DISCO}}$.}
  \label{fig:iam_heatmap}
\end{figure}

\subsubsection{Multilingual scene text (ICDAR$_{\text{DISCO}}$)}
\label{sec:parsing-icdar}

ICDAR$_{\text{DISCO}}$ spans 10 languages (Arabic, Bangla, Chinese, Hindi, Japanese, Korean, Italian, French, German) with 50 samples each. OCR systems struggle on non-Latin scripts, achieving $\SCORE{CER} > 5.0$, while VLMs maintain $\SCORE{CER} < 2.5$ across all scripts.

\begin{figure}[h]
  \centering
  \includegraphics[width=\textwidth]{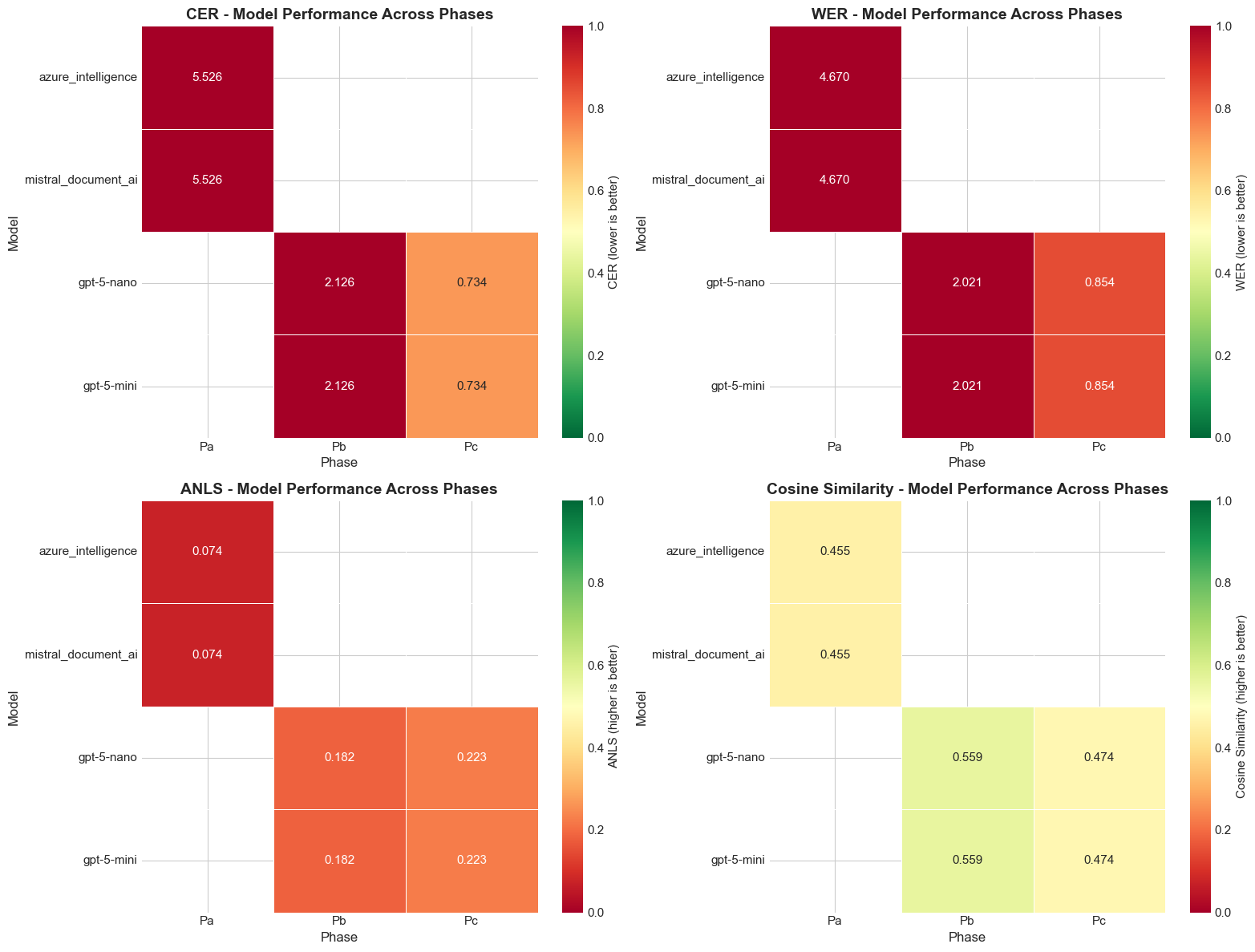}
  \caption{Model performance across phases on ICDAR$_{\text{DISCO}}$. $P_{\text{OCR}}$: OCR baseline; $P_{\text{VLM-base}}$: VLM with generic prompting; $P_{\text{VLM-task}}$: VLM with task-aware prompting. For CER and WER, lower (green) is better; for ANLS and Cosine Similarity, higher is better.}
  \label{fig:icdar_heatmap}
\end{figure}

\textbf{Results} Table~\ref{tab:parsing_overall} presents aggregate performance across all phases. VLMs consistently outperform dedicated OCR services, with task-aware prompting yielding the lowest error rates. The best configuration (VLM + context) achieves a mean $\SCORE{CER}$ of 0.73 compared to 5.53 for OCR baselines---a reduction of approximately 87\%. Beyond absolute performance, VLMs exhibit substantially lower variance ($\SCORE{CER}$ standard deviation of 0.40 vs 36.55), indicating more consistent behaviour across diverse scripts. Fig.~\ref{fig:icdar_heatmap} illustrates these patterns: OCR models cluster in the high-error region (dark red) while VLMs with task-aware prompting ($P_{\text{VLM-task}}$) achieve the lowest $\SCORE{CER}$ and $\SCORE{WER}$.

\begin{table}[h]
  \centering
  \small
  \begin{tabular}{lcccc}
    \toprule
    \textbf{Approach} & \textbf{CER} $\downarrow$ & \textbf{WER} $\downarrow$ & \textbf{ANLS} $\uparrow$ & \textbf{$\SCORE{CS}$} $\uparrow$ \\
    \midrule
    OCR models & 5.53  & 4.67 & 0.07 & 0.45 \\
    VLM (generic prompt) & 2.13 & 2.02  & 0.18 & 0.56 \\
    VLM (task-aware prompt) & \textbf{0.73}  & \textbf{0.85}  & \textbf{0.22} & 0.47 \\
    \bottomrule
  \end{tabular}
  \caption{Parsing average performance on ICDAR Mini. Results averaged across models within each phase.}
  \label{tab:parsing_overall}
\end{table}

\textbf{Script-level analysis.} Performance gaps widen for non-Latin scripts. Table~\ref{tab:parsing_scripts} shows selected language categories where VLMs demonstrate the largest improvements. For Chinese text, VLMs reduce CER from 5.04 to 1.93. The contrast is most pronounced for mixed-script content: OCR achieves a CER of 155.84 on ``Chinese, Mixed'' samples, while VLMs maintain a CER below 1.0. Similar patterns emerge for Hindi (CER 3.16 $\rightarrow$ 1.17) and Bangla (CER 2.45 $\rightarrow$ 3.35, though both struggle here).

\begin{table}[h]
  \centering
  \small
  \begin{tabular}{lccc}
    \toprule
    \textbf{Script Category} & \textbf{OCR (CER)} $\downarrow$ & \textbf{VLM (CER)} $\downarrow$ & \textbf{$n$} \\
    \midrule
    Chinese & 5.04 & \textbf{1.93} & 94 \\
    Chinese, Latin & 1.59 & \textbf{0.59} & 85 \\
    Hindi & 3.16 & \textbf{1.17} & 106 \\
    Japanese & 1.75 & \textbf{1.61} & 64 \\
    Korean & 0.89 & \textbf{0.35} & 67 \\
    Latin & \textbf{1.69} & 2.07 & 24 \\
    \bottomrule
  \end{tabular}
  \caption{CER by script category for best-performing OCR (\texttt{azure-ai-documentintelligence}) vs best-performing VLM (\texttt{gpt-5-mini}). Sample counts in parentheses.}
  \label{tab:parsing_scripts}
\end{table}

\textbf{Discussion.} These results suggest that VLMs offer a more robust solution for multilingual document parsing than specialised OCR services, particularly for non-Latin scripts and mixed-language content. The substantial benefit of task-aware prompting (66\% CER reduction over generic prompts) highlights the importance of prompt design in document intelligence applications. As shown in Fig.~\ref{fig:icdar_heatmap}, the cosine similarity for task-aware VLMs drops slightly (0.47 vs 0.56 for generic prompts), which may indicate semantic drift when prompts become overly prescriptive---though ANLS, a stricter matching metric, continues to improve. The VLM advantage is largest for Arabic ($+36.8\%$) and Bangla ($+40.2\%$), suggesting OCR systems were primarily optimised for Latin scripts while VLMs benefit from multilingual pretraining corpora.

\subsubsection{Medical documents (RxPad)}

\begin{figure}[h]
  \centering
  \includegraphics[width=\textwidth]{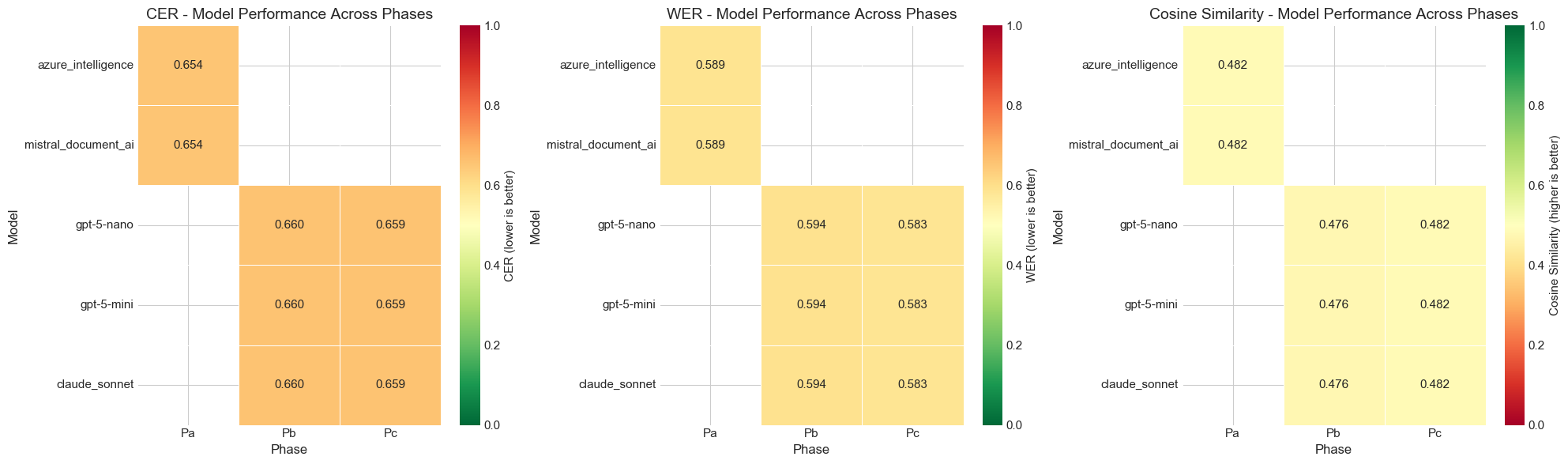}
  \caption{Model performance across phases on RxPad. $P_{\text{OCR}}$: OCR baseline; $P_{\text{VLM-base}}$: VLM with base prompting; $P_{\text{VLM-task}}$: VLM with task-aware prompting. For $\SCORE{CER}$ and $\SCORE{WER}$, lower (green) is better; for $\SCORE{CS}$, higher is better.}
  \label{fig:rxpad_heatmap}
\end{figure}

\textbf{Results} Table~\ref{tab:rxpad-metrics} summarises the performance of OCR and VLM approaches across all three experimental phases. Traditional OCR systems (\texttt{azure-ai-documentintelligence} and \texttt{mistral-ocr-2505}) achieved a mean $\SCORE{CER}$ of 0.654 and $\SCORE{WER}$ of 0.589 in Phase $P_{\text{OCR}}$. VLMs tested with base prompting (Phase $P_{\text{VLM-base}}$) performed comparably, with $\SCORE{CER}$ values of 0.660 and $\SCORE{WER}$ of 0.594. When provided with a task-aware medical prompt (Phase $P_{\text{VLM-task}}$), VLMs showed marginal improvement, reducing $\SCORE{WER}$ to 0.583 whilst maintaining similar $\SCORE{CER}$ (0.659). $\SCORE{CS}$ remained consistent across all approaches, ranging from 0.476 to 0.482. These patterns are clearly visible in Fig.~\ref{fig:rxpad_heatmap}, where the colour gradient reveals minimal variation between OCR and VLM performance. $\SCORE{ANLS}$ was near zero across all experiments; this metric is designed for exact-match question answering scenarios and is not suitable for this task, where output formatting differences between predictions and ground truth dominate the error signal.

\begin{table}[ht]
  \centering
  \begin{tabular}{lccc}
    \toprule
    \textbf{Phase} & \textbf{$\SCORE{CER}$} $\downarrow$ & \textbf{$\SCORE{WER}$} $\downarrow$ & \textbf{$\SCORE{CS}$} $\uparrow$ \\
    \midrule
    $P_{\text{OCR}}$ & 0.654 & 0.589 & 0.482 \\
    $P_{\text{VLM-base}}$ & 0.660 & 0.594 & 0.476 \\
    $P_{\text{VLM-task}}$ & 0.659 & 0.583 & 0.482 \\
    \bottomrule
  \end{tabular}
  \caption{Mean performance metrics across experimental phases on RxPad, averaged across all evaluated models.}
  \label{tab:rxpad-metrics}
\end{table}

\textbf{Discussion} The results indicate that neither OCR nor VLMs hold a clear advantage for raw text extraction on French medical prescriptions. As shown in Fig.~\ref{fig:rxpad_heatmap}, performance differences between approaches are marginal across all metrics. However, qualitative analysis of model outputs reveals an important distinction: VLMs consistently produce structured key-value representations (e.g., \texttt{product\_name: DOLIPRANE}, \texttt{dose\_unit: comprimé}), whilst OCR systems and ground truth annotations contain unstructured plain text. This format mismatch artificially inflates character and word error rates, as the models are penalised for reformatting rather than misunderstanding content.

Field-level extraction analysis (Table~\ref{tab:rxpad-fields}) supports this interpretation. VLMs achieved high recall on medical terminology such as medication names, dosage units, and prescription identifiers, suggesting that comprehension is not the limiting factor. The modest improvement observed when adding medical context ($P_{\text{VLM-task}}$ vs $P_{\text{VLM-base}}$) indicates that domain-specific prompting provides limited benefit for this dataset, likely because the visual and textual cues in prescription documents are already sufficient for general-purpose models to infer the clinical context.

\begin{table}[ht]
  \centering
  \begin{tabular}{lcc}
    \toprule
    \textbf{Field (French)} & \textbf{GT Presence} $\uparrow$ & \textbf{VLM Recall} $\uparrow$ \\
    \midrule
    mg (Milligrams) & 70.0\% & 100.0\% \\
    comprimé (Tablet) & 46.5\% & 100.0\% \\
    traitement (Treatment) & 31.5\% & 100.0\% \\
    médecin (Doctor) & 32.0\% & 139.1\% \\
    ordonnance (Prescription) & 13.5\% & 100.0\% \\
    fois par jour (Times/day) & 25.0\% & 4.0\% \\
    \bottomrule
  \end{tabular}
  \caption{Field extraction recall for VLMs in Phase $P_{\text{VLM-task}}$. Values above 100\% indicate the model detected more instances than present in ground truth annotations.}
  \label{tab:rxpad-fields}
\end{table}

These findings suggest that for clinical document processing pipelines, the choice between OCR and VLMs should be guided by downstream task requirements rather than raw extraction accuracy. VLMs may be preferable when structured output is desired, whilst OCR remains suitable for applications requiring verbatim text reproduction.

\clearpage
%==============================================================================
\subsection{QA task: in-depth analysis}
\label{appendix:qa_analysis}

\subsubsection{Document questions (DocVQA$_{\text{DISCO}}$)}

We evaluated three document QA strategies on the DocVQA benchmark (500 samples): OCR-based pipelines ($\QA{OCR}{strat}$), VLM parse-then-answer ($\QA{VLM-2stage}{strat}$), and direct visual question answering ($\QA{VLM-direct}{strat}$). Table~\ref{tab:docvqa_strategy} summarises the best-performing configuration within each strategy.

\begin{table}[h]
  \centering
  \begin{tabular}{llcccc}
    \toprule
    \textbf{Strategy} & \textbf{Best Configuration} & \textbf{$\SCORE{GT\text{-}in\text{-}Pred}$} $\uparrow$ & \textbf{$\SCORE{ANLS}$} $\uparrow$ & \textbf{$\SCORE{EM}$} $\uparrow$ & \textbf{$\SCORE{CS}$} $\uparrow$ \\
    \midrule
    $QA_{\text{OCR}}$ & \texttt{azure-ai-documentintelligence} $\rightarrow$ \texttt{gpt-5-mini} & 0.876 & 0.841 & 0.744 & 0.855 \\
    $QA_{\text{VLM-2stage}}$ & \texttt{gpt-5-mini} $\rightarrow$ \texttt{gpt-5-mini} & 0.896 & 0.868 & 0.779 & 0.876 \\
    $QA_{\text{VLM-direct}}$ & \texttt{gpt-5-mini} & \textbf{0.908} & 0.632 & 0.482 & 0.786 \\
    \bottomrule
  \end{tabular}
  \caption{Best model performance by strategy on DocVQA. $\SCORE{GT\text{-}in\text{-}Pred}$ is the primary metric.}
  \label{tab:docvqa_strategy}
\end{table}

\begin{figure}[h]
  \centering
  \includegraphics[width=0.5\textwidth]{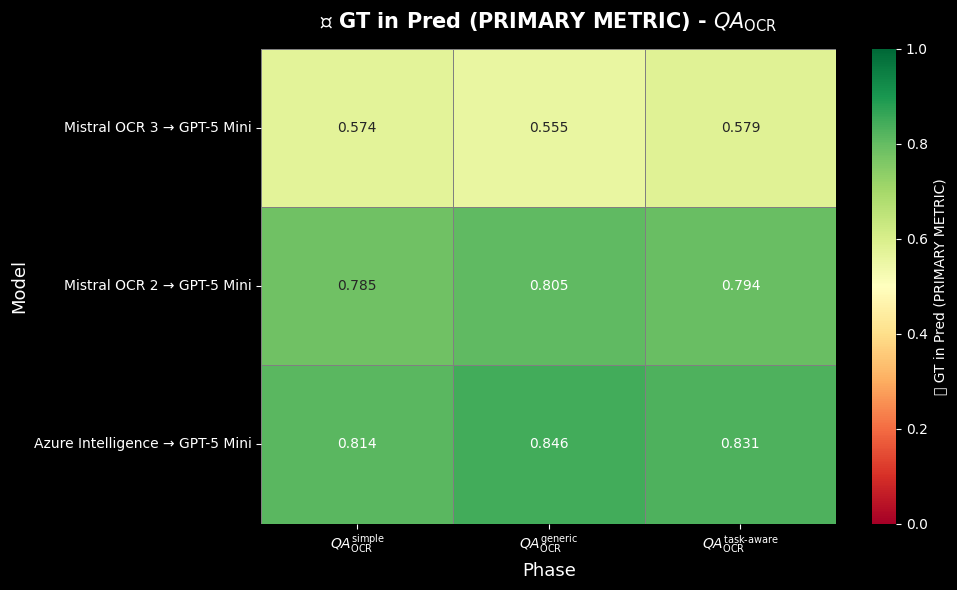}\\[0.5em]
  \includegraphics[width=0.5\textwidth]{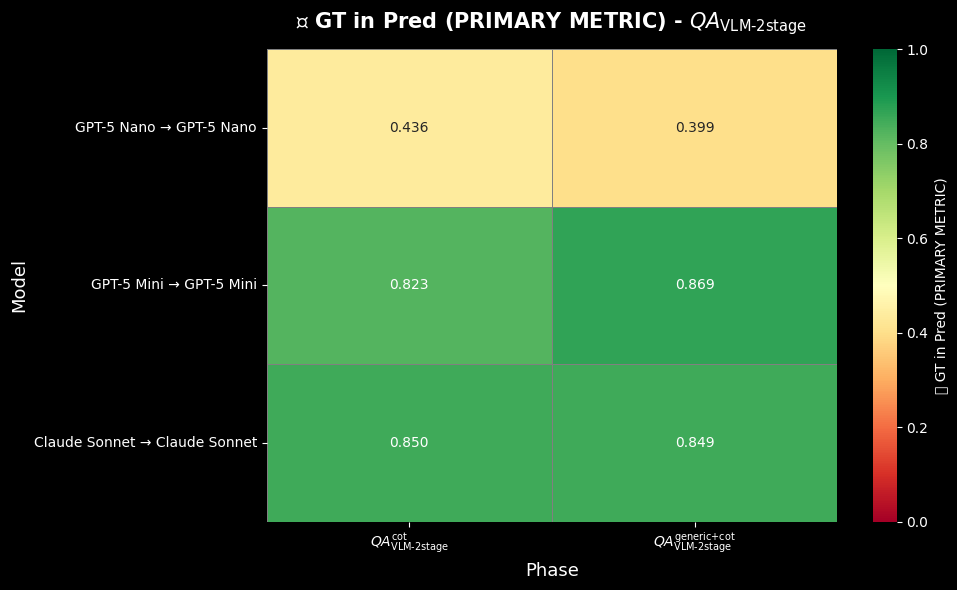}\\[0.5em]
  \includegraphics[width=0.5\textwidth]{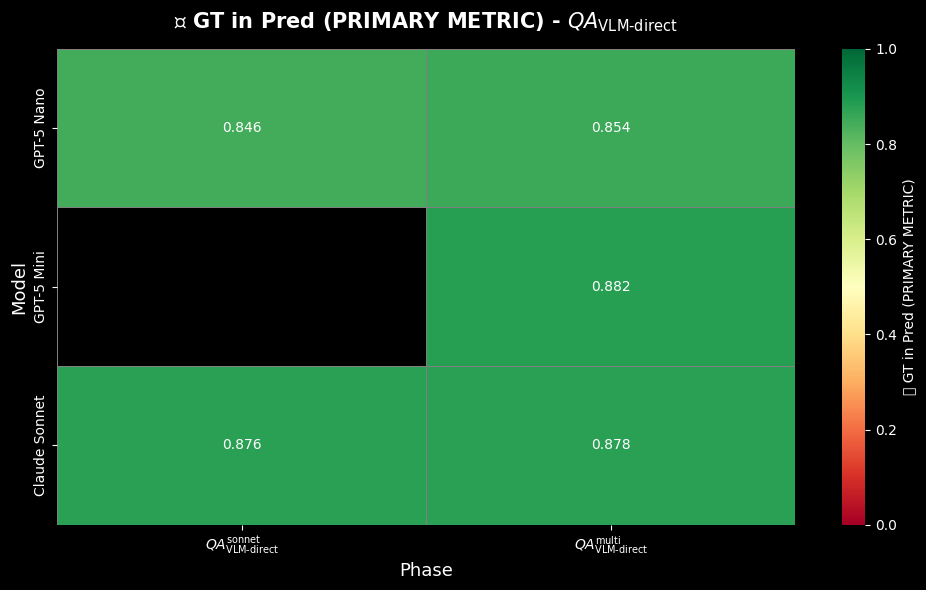}
  \caption{DocVQA strategy heatmaps using the primary strategy metric $\SCORE{GT\text{-}in\text{-}Pred}$ for (1) $QA_{\text{OCR}}$, (2) $QA_{\text{VLM-2stage}}$, and (3) $QA_{\text{VLM-direct}}$.}
  \label{fig:docvqa_heatmap}
\end{figure}

Direct VQA achieved the highest $\SCORE{GT\text{-}in\text{-}Pred}$ score (0.908), outperforming the best OCR-based pipeline by 3.2 percentage points. As shown in Fig.~\ref{fig:docvqa_heatmap}, this pattern holds consistently: all models evaluated under $QA_{\text{VLM-direct}}$ outperform their $QA_{\text{OCR}}$ and $QA_{\text{VLM-2stage}}$ counterparts on $\SCORE{GT\text{-}in\text{-}Pred}$. \texttt{gpt-5-nano}, for instance, achieves 0.877 with direct VQA compared to 0.425 when used in a parse-then-answer configuration.

\begin{figure}[h]
  \centering
  \includegraphics[width=\textwidth]{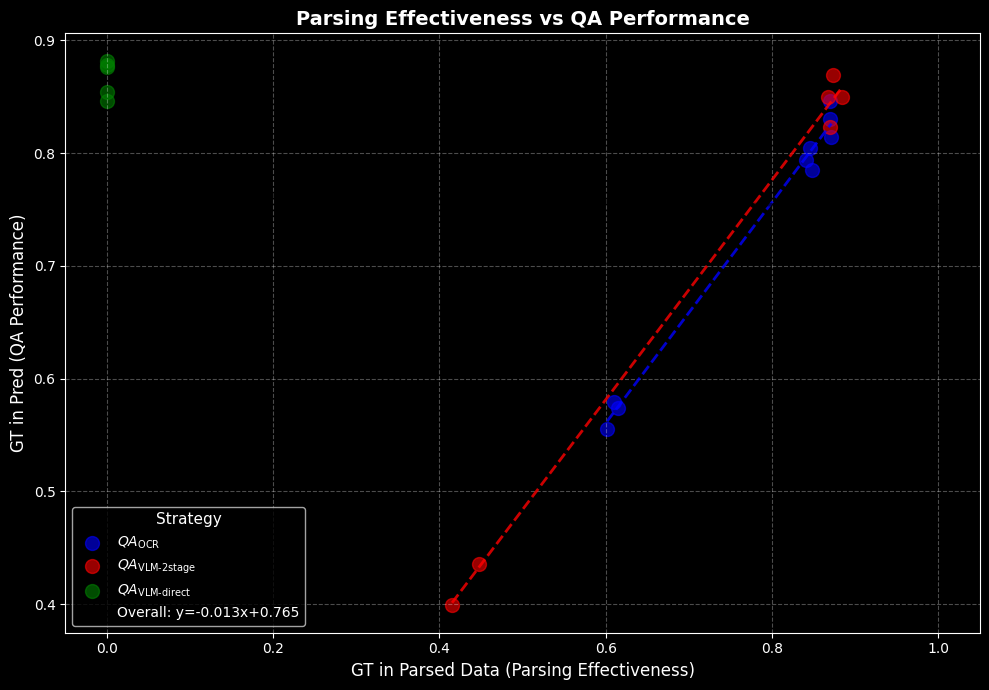}
  \caption{Regression performance when predicting QA correctness from the parsed document data (DocVQA).}
  \label{fig:docvqa-correctness-from-parsed-regression}
\end{figure}

However, the relationship between strategies reverses when considering string-matching metrics. Table~\ref{tab:docvqa-metrics-gap} quantifies the discrepancy between GT in Pred and ANLS across strategies.

\begin{table}[h]
  \centering
  \begin{tabular}{lcc}
    \toprule
    \textbf{Strategy} & \textbf{Mean $\SCORE{GT\text{-}in\text{-}Pred}$} $\uparrow$ & \textbf{Mean $\SCORE{ANLS}$} $\uparrow$ \\
    \midrule
    $QA_{\text{OCR}}$ & 0.840 & 0.720 \\
    $QA_{\text{VLM-2stage}}$ & 0.725 & 0.480 \\
    $QA_{\text{VLM-direct}}$ & 0.893 & 0.450 \\
    \bottomrule
  \end{tabular}
  \caption{Mean QA performance by strategy on DocVQA (reported separately for $\SCORE{GT\text{-}in\text{-}Pred}$ and $\SCORE{ANLS}$).}
  \label{tab:docvqa-metrics-gap}
\end{table}

The gap between $\SCORE{GT\text{-}in\text{-}Pred}$ and $\SCORE{ANLS}$ widens substantially for direct VQA, driven primarily by \texttt{claude-3-5-sonnet}'s behaviour. Fig.~\ref{fig:docvqa_heatmap} reveals that \texttt{claude-3-5-sonnet} achieves 0.904 $\SCORE{GT\text{-}in\text{-}Pred}$ but only 0.066 $\SCORE{ANLS}$ and 0.047 $\SCORE{EM}$ in direct VQA mode. This indicates that while the ground truth answer is present within \texttt{claude-3-5-sonnet}'s responses, the output format diverges significantly from the expected terse answers.

Within the OCR pipeline ($\QA{OCR}{strat}$), OCR provider selection has measurable impact. \texttt{azure-ai-documentintelligence} consistently outperforms \texttt{mistral-ocr-2505} by 3--4 percentage points on $\SCORE{GT\text{-}in\text{-}Pred}$ when paired with the same downstream QA model (0.876 vs 0.833 in $\QA{OCR}{cot}$).

\textbf{Discussion} The results suggest that for document QA, direct visual processing by VLMs outperforms explicit text extraction pipelines when measured by answer containment ($\SCORE{GT\text{-}in\text{-}Pred}$). This finding challenges the conventional assumption that OCR-based approaches provide superior text grounding. The VLM's ability to jointly reason over visual layout, typography, and textual content appears to confer an advantage over pipelines that discard spatial information during OCR.

\begin{table}[h]
  \centering
  \begin{tabular}{lc}
    \toprule
    \textbf{Model / system} & \textbf{$\SCORE{GT-in-Extracted-Text}$} $\uparrow$ \\
    \midrule
    \texttt{azure-ai-documentintelligence} & 0.8712 \\
    \texttt{mistral-ocr-2505} & 0.8479 \\
    \texttt{mistral-ocr-2512} & 0.6146 \\
    \texttt{gpt-5-nano} & 0.4479 \\
    \texttt{gpt-5-mini} & 0.8695 \\
    \texttt{claude-3.5-sonnet} & 0.8677 \\
    \bottomrule
  \end{tabular}
  \caption{\textbf{$\SCORE{GT-in-Extracted-Text}$} from parsed document data on DocVQA.}
  \label{tab:docvqa_gt_in_parsed}
\end{table}

The divergence between $\SCORE{GT\text{-}in\text{-}Pred}$ and string-matching metrics warrants careful interpretation. High $\SCORE{GT\text{-}in\text{-}Pred}$ with low $\SCORE{ANLS}$ indicates verbose but correct responses---the model identifies the right information but embeds it within explanatory text. This behaviour is particularly pronounced in \texttt{\texttt{claude-3-5-sonnet}} across all strategies, as evidenced in Fig.~\ref{fig:docvqa_heatmap}. Whether this constitutes a limitation depends on downstream requirements: extractive applications requiring structured outputs would penalise such responses, whereas information retrieval or human-facing systems may prefer them.

The poor performance of VLM parse-then-answer ($\QA{VLM-2stage}{strat}$) relative to both alternatives is notable. Despite using the same model for both stages, \texttt{gpt-5-nano} achieves only 0.425 $\SCORE{GT\text{-}in\text{-}Pred}$ in $\QA{VLM-2stage}{strat}$ versus 0.877 in $\QA{VLM-direct}{strat}$. This suggests that the intermediate text representation introduces information loss or formatting artifacts that degrade downstream QA performance, without the compensating benefit of specialised OCR systems used in $\QA{OCR}{strat}$.

OCR quality remains relevant within two-stage pipelines. The consistent 3--4 point advantage of \texttt{azure-ai-documentintelligence} over \texttt{mistral-ocr-2505} indicates that OCR errors propagate to QA performance, supporting the intuition that text extraction fidelity bounds downstream accuracy in pipeline architectures.Also, table \ref{tab:docvqa_gt_in_parsed} and figure {fig:docvqa-correctness-from-parsed-regression} both confirm that these results are mainly error propagation from parsing tasks and not only due to the QA stage.

Within the OCR pipeline strategy, provider choice significantly impacts performance on structured single-page documents. Table~\ref{tab:docvqa_ocr_detailed} presents phase-wise performance breakdown.

\begin{table}[h]
  \centering
  \begin{tabular}{l|ccc}
    \toprule
    \textbf{OCR system} & $\QA{OCR}{generic}$ & $\QA{OCR}{cot}$ & $\QA{OCR}{task-aware}$ \\
    \midrule
    \multicolumn{4}{l}{\textit{$\SCORE{GT-in-Pred}$ (primary metric) $\uparrow$}} \\
    \texttt{azure-ai-documentintelligence} & 0.814 & \textbf{0.846} & 0.831 \\
    \texttt{mistral-ocr-2505} & 0.785 & 0.805 & 0.794 \\
    \texttt{mistral-ocr-2512} & 0.574 & 0.555 & 0.579 \\
    \midrule
    \multicolumn{4}{l}{\textit{$\SCORE{ANLS}$ (string similarity) $\uparrow$}} \\
    \texttt{azure-ai-documentintelligence} & 0.623 & \textbf{0.833} & 0.733 \\
    \texttt{mistral-ocr-2505} & 0.607 & 0.787 & 0.692 \\
    \texttt{mistral-ocr-2512} & 0.459 & 0.560 & 0.499 \\
    \midrule
    \multicolumn{4}{l}{\textit{$\SCORE{EM}$ (exact match) $\uparrow$}} \\
    \texttt{azure-ai-documentintelligence} & 0.454 & \textbf{0.720} & 0.559 \\
    \texttt{mistral-ocr-2505} & 0.444 & 0.696 & 0.521 \\
    \texttt{mistral-ocr-2512} & 0.331 & 0.507 & 0.374 \\
    \bottomrule
  \end{tabular}
  \caption{OCR system performance on DocVQA$_{\text{DISCO}}$ by phase (all using \texttt{gpt-5-mini} for QA).}
  \label{tab:docvqa_ocr_detailed}
\end{table}

Chain-of-thought prompting ($QA_{\text{OCR}}^{\text{cot}}$) yielded the strongest performance across all OCR systems, with Azure Intelligence achieving 0.846 $\SCORE{GT-in-Pred}$ and 0.720 exact match rate. Notably, the performance ordering remained consistent across prompting strategies: Azure > Mistral OCR 2 > Mistral OCR 3. The 26-point gap between Mistral OCR versions (0.805 vs 0.555 in $QA_{\text{OCR}}^{\text{cot}}$) exceeded the gap between Azure and Mistral OCR 2 (4.1 points), indicating that Mistral OCR 3 represents a substantial regression rather than incremental improvement.

\subsubsection{InfographicVQA}

\textbf{Results} We evaluated three document QA strategies on the InfographicVQA\_mini benchmark (500 infographic question-answer pairs): (i) OCR+VLM pipelines where a dedicated OCR system extracts text before an LLM answers the question ($QA_{\text{OCR}}$), (ii) VLM Parse+QA where the same vision-language model performs both parsing and answering ($QA_{\text{VLM-2stage}}$), and (iii) Direct VQA where the VLM receives the image and question without intermediate text extraction ($QA_{\text{VLM-direct}}$). Performance was measured using $\SCORE{GT-in-Pred}$, $\SCORE{ANLS}$, and $\SCORE{EM}$.

Table~\ref{tab:strategy_comparison} presents the best-performing configuration for each strategy using \texttt{gpt-5-mini} as the QA model. Direct VQA achieved the highest $\SCORE{GT\text{-}in\text{-}Pred}$ (0.785), indicating that the correct answer was contained in the model's response more frequently than with other approaches. However, it exhibited substantially lower $\SCORE{ANLS}$ (0.186) and $\SCORE{EM}$ (0.102), suggesting verbose or poorly formatted outputs rather than incorrect answers. The OCR+VLM pipeline using \texttt{azure-ai-documentintelligence} achieved the most balanced performance across all metrics, with the highest $\SCORE{ANLS}$ (0.629) and $\SCORE{EM}$ (0.515).

\begin{figure*}[t]
  \centering

  \begin{minipage}[t]{0.95\textwidth}
    \centering
    \textbf{$QA_{\text{OCR}}$}\\
    \includegraphics[width=\linewidth]{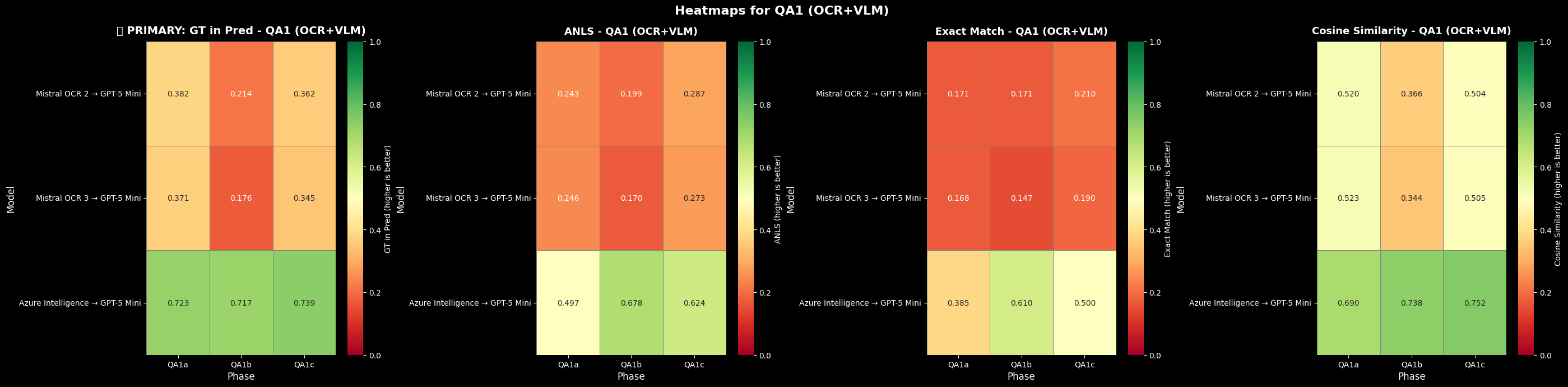}
  \end{minipage}\\[0.75em]

  \begin{minipage}[t]{0.95\textwidth}
    \centering
    \textbf{$QA_{\text{VLM-2stage}}$}\\
    \includegraphics[width=\linewidth]{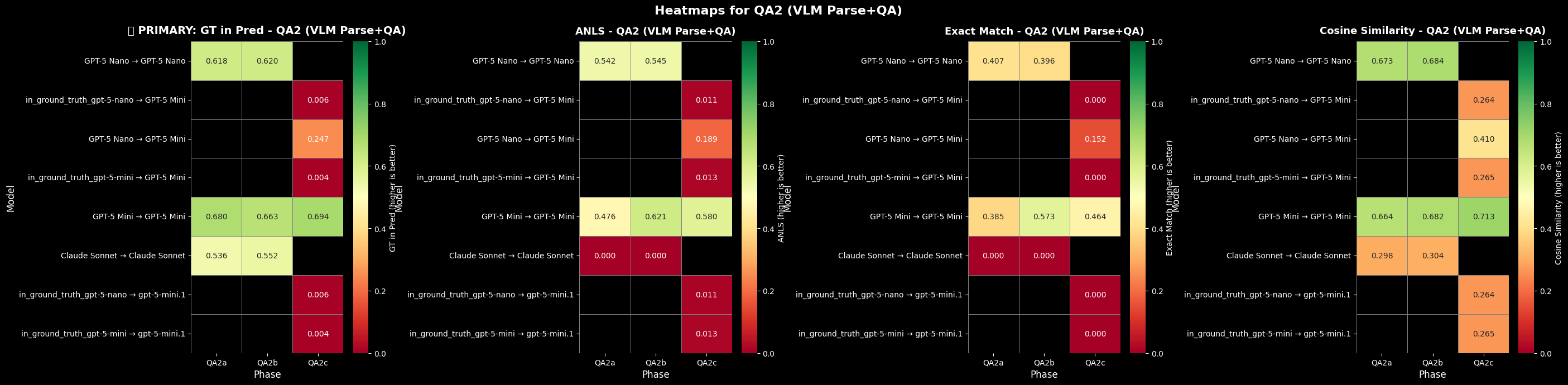}
  \end{minipage}\\[0.75em]

  \begin{minipage}[t]{0.95\textwidth}
    \centering
    \textbf{$QA_{\text{VLM-direct}}$}\\
    \includegraphics[width=\linewidth]{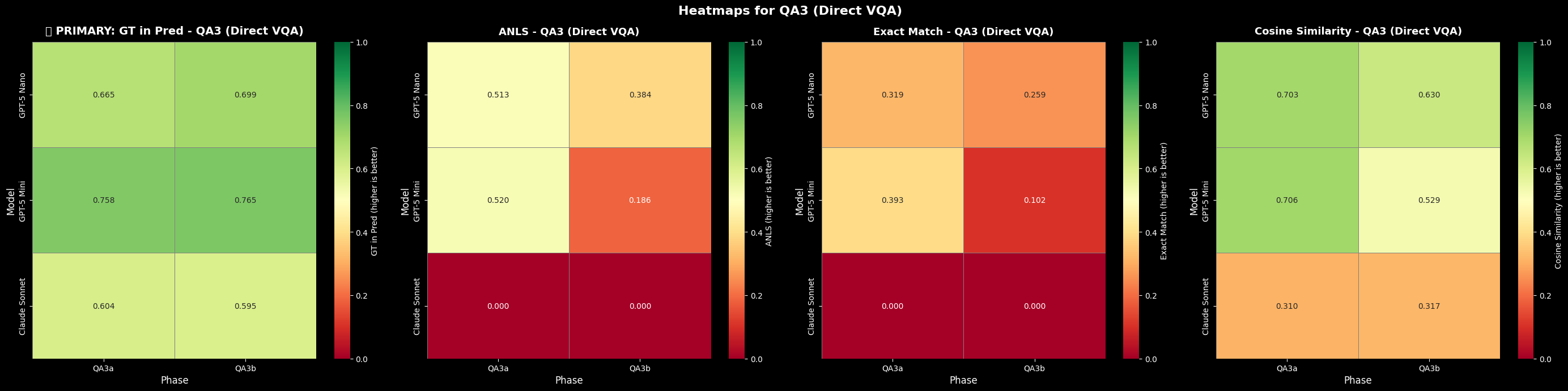}
  \end{minipage}

  \caption{InfographicVQA strategy heatmaps. Each row corresponds to a QA strategy ($QA_{\text{OCR}}$, $QA_{\text{VLM-2stage}}$, $QA_{\text{VLM-direct}}$) and reports four metrics: $\SCORE{GT\text{-}in\text{-}Pred}$, $\SCORE{ANLS}$, $\SCORE{EM}$, and $\SCORE{CS}$ (Cosine Similarity).}
  \label{fig:infovqa_strategy_heatmaps}
\end{figure*}

\begin{table}[h]
  \centering
  \begin{tabular}{lccc}
    \toprule
    \textbf{Strategy} & \textbf{$\SCORE{GT\text{-}in\text{-}Pred}$} $\uparrow$ & \textbf{$\SCORE{ANLS}$} $\uparrow$ & \textbf{$\SCORE{EM}$} $\uparrow$ \\
    \midrule
    $QA_{\text{OCR}}$: OCR$\to$QA (\texttt{azure-ai-documentintelligence}) & 0.754 & 0.629 & 0.515 \\
    $QA_{\text{VLM-2stage}}$: VLM$\to$QA & 0.711 & 0.585 & 0.477 \\
    $QA_{\text{VLM-direct}}$: Direct VQA & \textbf{0.785} & 0.186 & 0.102 \\
    \bottomrule
  \end{tabular}
  \caption{Best-performing configuration per strategy (\texttt{gpt-5-mini} as QA model)}
  \label{tab:strategy_comparison}
\end{table}

The choice of OCR system proved critical for pipeline-based approaches. As shown in Table~\ref{tab:ocr_comparison}, \texttt{azure-ai-documentintelligence} outperformed \texttt{mistral-ocr-2505} by a factor of two across all metrics, demonstrating that OCR quality represents a significant bottleneck in two-stage pipelines.

\begin{table}[h]
  \centering
  \begin{tabular}{lccc}
    \toprule
    \textbf{OCR System} & \textbf{$\SCORE{GT\text{-}in\text{-}Pred}$} $\uparrow$ & \textbf{$\SCORE{ANLS}$} $\uparrow$ & \textbf{$\SCORE{EM}$} $\uparrow$ \\
    \midrule
    \texttt{azure-ai-documentintelligence} & 0.754 & 0.629 & 0.515 \\
    \texttt{mistral-ocr-2505} & 0.368 & 0.288 & 0.214 \\
    \texttt{mistral-ocr-2512} & 0.345 & 0.273 & 0.190 \\
    \bottomrule
  \end{tabular}
  \caption{Impact of OCR system on $QA_{\text{OCR}}$ pipeline performance ($QA_{\text{OCR}}$-c phase, \texttt{gpt-5-mini})}
  \label{tab:ocr_comparison}
\end{table}

\textbf{Discussion} The results reveal a key distinction between answer correctness and answer format compliance. Direct VQA achieved the highest $\SCORE{GT\text{-}in\text{-}Pred}$, meaning VLMs can accurately locate and reason about the relevant information in infographics when given direct visual access. The discrepancy with $\SCORE{ANLS}$ and $\SCORE{EM}$ stems from response verbosity rather than factual errors---the model produces contextualised answers instead of terse extractions. This behaviour is addressable through prompt engineering: constraining the expected output format (e.g., instructing the model to respond with only the answer value) would likely align $\SCORE{ANLS}$ and $\SCORE{EM}$ with the $\SCORE{GT\text{-}in\text{-}Pred}$ performance.

OCR-based pipelines remain competitive when high-quality OCR is available, but their ceiling is fundamentally limited by text extraction fidelity. Infographics present particular challenges for OCR due to non-standard layouts, embedded text in visual elements, and the need to preserve spatial relationships between data points. Direct VQA circumvents these issues entirely by reasoning over the visual representation.

Given that (i) direct VQA demonstrates superior answer correctness as measured by $\SCORE{GT\text{-}in\text{-}Pred}$, (ii) the format compliance gap is a prompt-level rather than capability-level limitation, and (iii) end-to-end approaches avoid error propagation from OCR failures, we conclude that direct VQA with VLMs represents the most promising approach for infographic question answering. The observed verbosity is an engineering problem with known solutions, whereas OCR errors on complex visual documents remain a harder challenge.

\subsubsection{Multi-page documents (DUDE)}

\begin{figure*}[t]
  \centering

  \begin{minipage}[t]{0.95\textwidth}
    \centering
    \textbf{$QA_{\text{OCR}}$}\\
    \includegraphics[width=\linewidth]{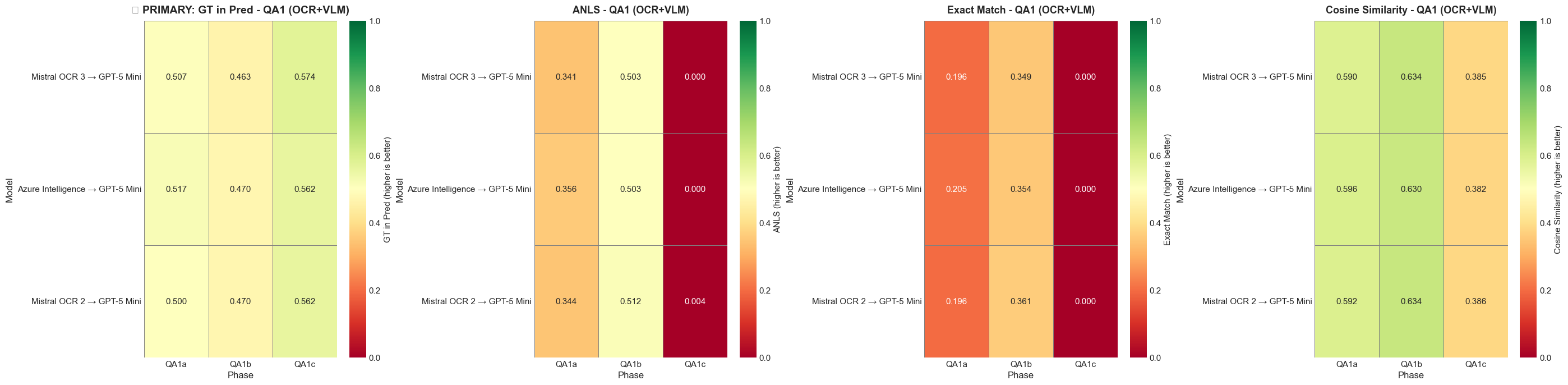}
  \end{minipage}\\[0.75em]

  \begin{minipage}[t]{0.95\textwidth}
    \centering
    \textbf{$QA_{\text{VLM-2stage}}$}\\
    \includegraphics[width=\linewidth]{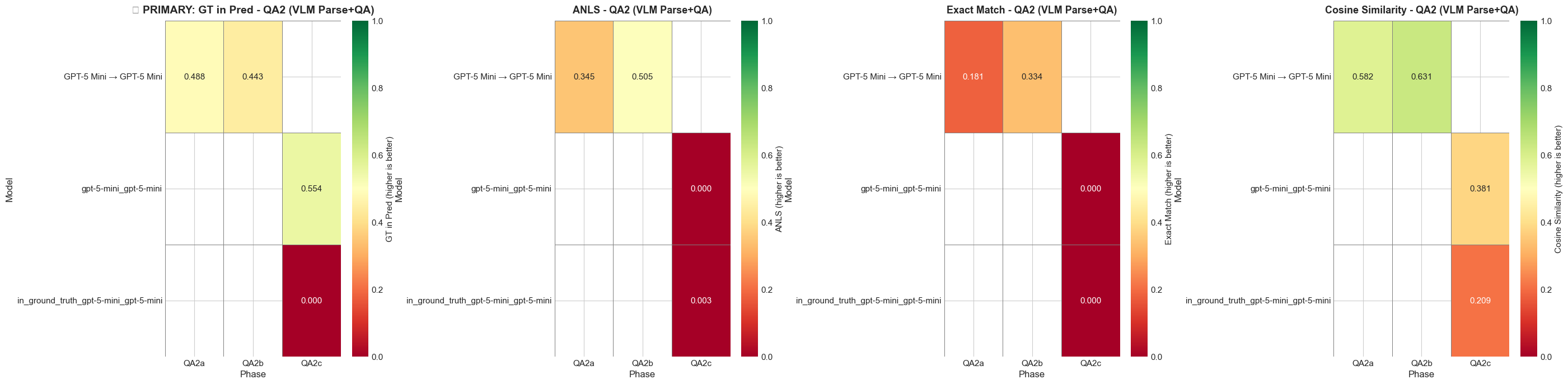}
  \end{minipage}\\[0.75em]

  \begin{minipage}[t]{0.95\textwidth}
    \centering
    \textbf{$QA_{\text{VLM-direct}}$}\\
    \includegraphics[width=\linewidth]{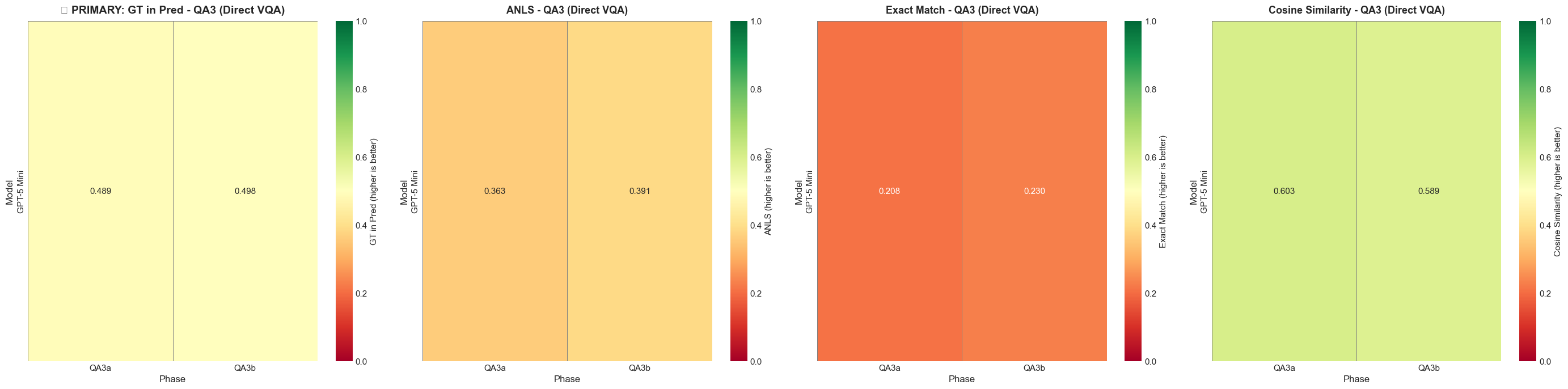}
  \end{minipage}

  \caption{DUDE strategy heatmaps. Each row corresponds to a QA strategy ($QA_{\text{OCR}}$, $QA_{\text{VLM-2stage}}$, $QA_{\text{VLM-direct}}$) and reports four metrics: $\SCORE{GT\text{-}in\text{-}Pred}$, $\SCORE{ANLS}$, $\SCORE{EM}$, and $\SCORE{CS}$ (Cosine Similarity).}
  \label{fig:dude_strategy_heatmaps}
\end{figure*}

\begin{table}[h]
  \centering
  \small
  \setlength{\tabcolsep}{8pt}
  \begin{tabular}{lc}
    \toprule
    \textbf{Model / parsing method} & \textbf{$\SCORE{GT\text{-}in\text{-}Parsed\text{-}Data}$} $\uparrow$ \\
    \midrule
    \texttt{azure-ai-documentintelligence} & \textbf{0.746} \\
    \texttt{mistral-ocr-2505}             & 0.357 \\
    \texttt{mistral-ocr-2512}             & 0.321 \\
    \midrule
    \texttt{gpt-5-nano}                   & 0.625 \\
    \texttt{gpt-5-mini}                   & 0.694 \\
    \texttt{claude-3-5-sonnet}            & 0.645 \\
    \bottomrule
  \end{tabular}
  \caption{Ground-truth coverage in parsed data across OCR-based and VLM-based parsing methods.}
  \label{tab:gt_parsed_extended}
\end{table}

Three strategies were compared:
\begin{itemize}
  \item \textbf{$\QA{OCR}{strat}$ (OCR+VLM)}: Dedicated OCR extraction (\texttt{azure-ai-documentintelligence} or \texttt{mistral-ocr-2505}) followed by \texttt{gpt-5-mini} for question answering over the extracted text.
  \item \textbf{$\QA{VLM-2stage}{strat}$ (VLM Parse+QA)}: \texttt{gpt-5-mini} performs both document parsing and subsequent question answering in a two-stage pipeline.
  \item \textbf{$\QA{VLM-direct}{strat}$ (Direct VQA)}: \texttt{gpt-5-mini} receives document images directly and answers questions without explicit text extraction.
\end{itemize}

The primary evaluation metric is $\SCORE{GT\text{-}in\text{-}Pred}$ (ground truth substring presence in prediction), supplemented by $\SCORE{ANLS}$ (Average Normalised Levenshtein Similarity), $\SCORE{EM}$, and substring match rates.

\textbf {Results} Table~\ref{tab:strategy} presents the aggregated performance across strategies. The hybrid OCR+VLM approach ($\QA{OCR}{strat}$) achieved the highest $\SCORE{GT\text{-}in\text{-}Pred}$ of 0.514, outperforming both direct VQA (0.493) and the VLM-based parsing pipeline (0.371). The performance gap between $\QA{OCR}{strat}$ and $\QA{VLM-2stage}{strat}$ is substantial (14.2 percentage points), indicating that VLM-based text extraction introduces errors that propagate to the QA stage.

Direct VQA ($\QA{VLM-direct}{strat}$) achieved the highest $\SCORE{ANLS}$ (0.377) and $\SCORE{EM}$ (21.9\%), suggesting that when answers are correct, they tend to be more precisely formatted. However, its lower $\SCORE{GT\text{-}in\text{-}Pred}$ indicates more frequent complete misses compared to the OCR-based approach.

\begin{table}[h]
  \centering
  \begin{tabular}{lcccc}
    \toprule
    \textbf{Strategy} & \textbf{$\SCORE{GT\text{-}in\text{-}Pred}$} $\uparrow$ & \textbf{$\SCORE{ANLS}$} $\uparrow$ & \textbf{$\SCORE{EM}$} $\uparrow$ & \textbf{Substring Match} $\uparrow$ \\
    \midrule
    $\QA{OCR}{strat}$ (OCR+VLM)        & \textbf{0.514} & 0.286 & 0.186 & \textbf{0.538} \\
    $\QA{VLM-2stage}{strat}$ (VLM Parse+QA)   & 0.371          & 0.213 & 0.129 & 0.391 \\
    $\QA{VLM-direct}{strat}$ (Direct VQA)     & 0.493          & \textbf{0.377} & \textbf{0.219} & 0.524 \\
    \bottomrule
  \end{tabular}
  \caption{Strategy-level performance comparison on DUDE (n=404 per phase)}
  \label{tab:strategy}
\end{table}

Table~\ref{tab:ocr} details the OCR tool comparison within the $\QA{OCR}{strat}$ strategy. \texttt{azure-ai-documentintelligence} and \texttt{mistral-ocr-2505} performed comparably, with differences of less than 2 percentage points on $\SCORE{GT\text{-}in\text{-}Pred}$ across phases. Both OCR systems showed identical performance on the $\QA{OCR}{cot}$ and $\QA{OCR}{task}$ phases ($\SCORE{GT\text{-}in\text{-}Pred}$ = 0.470 and 0.562 respectively), suggesting that downstream QA model behaviour dominates over OCR tool selection for these document types.

\begin{table}[h]
  \centering
  \begin{tabular}{llcccc}
    \toprule
    \textbf{Phase} & \textbf{OCR Tool} & \textbf{$\SCORE{GT\text{-}in\text{-}Pred}$} $\uparrow$ & \textbf{$\SCORE{ANLS}$} $\uparrow$ & \textbf{$\SCORE{EM}$} $\uparrow$ & \textbf{$\SCORE{CS}$} $\uparrow$ \\
    \midrule
    $\QA{OCR}{generic}$ & \texttt{azure-ai-documentintelligence} & 0.517 & 0.356 & 0.205 & 0.596 \\
    $\QA{OCR}{generic}$ & \texttt{mistral-ocr-2505}         & 0.500 & 0.344 & 0.196 & 0.592 \\
    $\QA{OCR}{generic}$ & \texttt{mistral-ocr-2512} & 0.512 & 0.341 & 0.196 & 0.591 \\
    \\
    \midrule
    $\QA{OCR}{cot}$ & \texttt{azure-ai-documentintelligence} & 0.470 & 0.503 & 0.354 & 0.630 \\
    $\QA{OCR}{cot}$ & \texttt{mistral-ocr-2505}         & 0.470 & 0.512 & 0.361 & 0.634 \\
    $\QA{OCR}{cot}$ & \texttt{mistral-ocr-2512} & 0.468 & 0.504 & 0.351 & 0.635 \\
    \midrule
    $\QA{OCR}{task}$ & \texttt{azure-ai-documentintelligence} & 0.562 & 0.000 & 0.000 & 0.382 \\
    $\QA{OCR}{task}$ & \texttt{mistral-ocr-2505}         & 0.562 & 0.004 & 0.000 & 0.386 \\
    $\QA{OCR}{task}$ & \texttt{mistral-ocr-2512} & 0.579 & 0.000 & 0.000 & 0.386 \\
    \bottomrule
  \end{tabular}
  \caption{OCR tool comparison within $\QA{OCR}{strat}$ (OCR $\rightarrow$ \texttt{gpt-5-mini})}
  \label{tab:ocr}
\end{table}

\textbf{Discussion} The results indicate that dedicated OCR remains advantageous over VLM-based parsing for document QA on complex multi-page documents. The $\QA{OCR}{strat}$ strategy's 14-point lead over $\QA{VLM-2stage}{strat}$ demonstrates that specialised OCR tools extract text more reliably than VLMs prompted to parse document content. This finding aligns with the architectural differences: OCR systems are explicitly trained for text localisation and recognition, whereas VLMs must jointly attend to visual layout and textual content.

The relatively strong performance of direct VQA ($\QA{VLM-direct}{strat}$) is noteworthy. By bypassing explicit text extraction, this approach avoids cascading OCR errors whilst retaining visual context. Its higher $\SCORE{ANLS}$ and $\SCORE{EM}$ suggest that when the model correctly identifies the answer location, it reproduces the text more faithfully than pipeline approaches. However, the 2-point $\SCORE{GT\text{-}in\text{-}Pred}$ deficit relative to $\QA{OCR}{strat}$ indicates that direct VQA more frequently fails to locate relevant information entirely.

The $\QA{VLM-2stage}{strat}$ strategy's poor performance ($\SCORE{GT\text{-}in\text{-}Pred}$ = 0.371) demonstrates the cost of error compounding in two-stage VLM pipelines. When the same model performs both parsing and QA, extraction errors in the first stage directly degrade QA accuracy. This suggests that if VLM-based parsing is required, using different models or architectures for each stage may mitigate error propagation.

The $\QA{OCR}{c}$ phase exhibits anomalous behaviour: high $\SCORE{GT\text{-}in\text{-}Pred}$ (0.562) but near-zero $\SCORE{ANLS}$ and $\SCORE{EM}$. This pattern indicates that predictions contain the ground truth as a substring but include substantial extraneous content, likely reflecting verbosity in the QA model's response format for this phase configuration.

The minimal performance difference between \texttt{azure-ai-documentintelligence} and \texttt{mistral-ocr-2505} (Table~\ref{tab:ocr}) suggests that for DUDE-style documents, the choice of OCR backend is less consequential than the overall pipeline architecture. Both commercial OCR systems achieve comparable text extraction quality on this benchmark.

\begin{table}[h]
  \centering
  \small
  \setlength{\tabcolsep}{8pt}
  \begin{tabular}{lc}
    \toprule
    \textbf{Parsing method} & \textbf{Ground truth in parsed data} \\
    \midrule
    Azure Intelligence OCR & 0.505 \\
    Mistral OCR 3          & 0.391 \\
    Mistral OCR 2          & 0.468 \\
    GPT-5 Mini             & 0.465 \\
    \bottomrule
  \end{tabular}
  \caption{Ground truth coverage in parsed data for different parsing methods.}
  \label{tab:gt_parsed_pipeline}
\end{table}

\textbf{Parsing effectiveness determines the QA ceiling on multi-page documents.} The relationship between OCR quality and QA performance is particularly clear on DUDE$_{\text{DISCO}}$, where ground-truth coverage in the parsed text predicts downstream accuracy. \texttt{azure-ai-documentintelligence} captured 50.5\% of answers in the extracted text, whereas \texttt{mistral-ocr-2512} captured only 39.1\%---an 11.4 percentage point gap that translated into similar differences in final QA performance.

\begin{table}[h]
  \centering
  \begin{tabular}{l|cc|cc}
    \toprule
    & \multicolumn{2}{c|}{Parsing} & \multicolumn{2}{c}{QA Performance} \\
    OCR System & GT-in-Ext & Rank & QA$^{\text{cot}}_{\text{OCR}}$ & QA$^{\text{task}}_{\text{OCR}}$ \\
    \midrule
    Azure Intelligence & 0.505 & 1 & 0.475 (2) & 0.567 (2) \\
    Mistral OCR 2 & 0.468 & 2 & \textbf{0.475} (2) & \textbf{0.567} (2) \\
    Mistral OCR 3 & 0.391 & 3 & 0.468 (3) & 0.579 (1) \\
    \bottomrule
  \end{tabular}
  \caption{Relationship between parsing effectiveness (GT-in-Extracted-Text) and QA performance on DUDEDISCO.}
  \label{tab:dude-parsing-vs-qa}
\end{table}

Interestingly, QA performance rankings did not perfectly mirror parsing effectiveness. On QA$^{\text{task}}_{\text{OCR}}$, Mistral OCR 3 achieved the highest GT-in-Pred (0.579) despite weakest parsing (0.391), suggesting that downstream prompt engineering can partially compensate for poor text extraction—though at the cost of format compliance (ANLS = 0.000).

%==============================================================================
\subsubsection{OCR System Comparison Across Datasets}
% \label{sec:ocr-ceiling}

% \subsubsection{Parsing quality establishes a ceiling}
\label{sec:ocr-ceiling}
On multi-page documents, downstream QA performance is bounded by parsing/coverage: missing text cannot be recovered by the QA model, so improvements in OCR/text extraction reliability translate into measurable QA gains.

Within OCR pipelines, provider choice has dataset-dependent effects. On DUDE$_{\text{DISCO}}$, \texttt{azure-ai-documentintelligence} and \texttt{mistral-ocr-2505} performed identically under chain-of-thought prompting ($\QA{OCR}{cot}$; $\SCORE{GT\text{-}in\text{-}Pred}=0.470$ for both), suggesting comparable text extraction quality. However, on DocVQA$_{\text{DISCO}}$, \texttt{azure-ai-documentintelligence} outperformed \texttt{mistral-ocr-2505} by 3.3 percentage points (0.876 vs 0.843), indicating that single-page form documents benefit from Azure's stronger layout analysis. \texttt{mistral-ocr-2512} (a newer model) underperformed consistently across datasets, suggesting that version number alone does not guarantee improved document understanding.
%==============================================================================

\textbf{Mistral OCR 3 underperforms its predecessor across all benchmarks.} Table~\ref{tab:mistral-ocr-comparison} shows that despite being the newer model version (mistral-ocr-2512 vs mistral-ocr-2505), Mistral OCR 3 consistently achieves lower parsing effectiveness than Mistral OCR 2. On DocVQADISCO, Mistral OCR 3 captured only 61.5\% of ground-truth answers in extracted text compared to 84.8\% for Mistral OCR 2—a 23.3 percentage point degradation. This pattern holds across all three QA datasets, challenging the assumption that newer model versions automatically deliver improved performance.

\begin{table}[h]
  \centering
  \begin{tabular}{l|ccc}
    \toprule
    OCR System & DocVQA & InfographicVQA & DUDE \\
    \midrule
    Azure Intelligence & \textbf{0.871} & \textbf{0.754} & \textbf{0.505} \\
    Mistral OCR 2 (2505) & 0.848 & 0.357 & 0.468 \\
    Mistral OCR 3 (2512) & 0.615 & 0.321 & 0.391 \\
    \midrule
    \multicolumn{4}{l}{\textit{GPT-5 Mini (VLM parsing for reference)}} \\
    GPT-5 Mini & 0.869 & 0.465 & 0.694 \\
    \bottomrule
  \end{tabular}
  \caption{Ground truth coverage in parsed text by OCR system (GT-in-Extracted-Text metric). Higher values indicate more reliable text extraction.}
  \label{tab:mistral-ocr-comparison}
\end{table}

\textbf{The performance gap translates directly to downstream QA accuracy.} On DocVQA$_{\text{DISCO}}$, the 23-point parsing disadvantage of \texttt{mistral-ocr-2512} (relative to \texttt{mistral-ocr-2505}) resulted in a 26-point drop in QA performance in the $\QA{OCR}{cot}$  (0.555 vs 0.805).

\textbf{Azure Intelligence maintains a consistent advantage on structured documents.} Azure OCR outperformed both Mistral systems on DocVQA$_{\text{DISCO}}$ (forms, letters) and InfographicVQA$_{\text{DISCO}}$ (visual layouts), likely due to superior layout analysis capabilities. However, on DUDE$_{\text{DISCO}}$ (multi-page documents), the gap narrowed substantially, with Mistral OCR 2 achieving comparable performance (0.468 vs 0.505). This suggests that layout understanding matters more for single-page structured documents than for multi-page, text-heavy content.

\textbf{\QApipe{1}{c} phase reveals systematic verbosity issues.} Table~\ref{tab:qa1c_anomaly} shows that all OCR systems achieved high GT-in-Pred scores in \QApipe{1}{c} (DUDE$_{\text{DISCO}}$: 0.562--0.579) but near-zero ANLS and EM. This pattern---correct answer present but format non-compliant---indicates that the task-aware prompt encouraged verbose responses. Mistral OCR 3 showed slightly higher GT-in-Pred (0.579 vs 0.567 for Azure), suggesting better answer localisation despite weaker parsing, but this advantage was negated by formatting issues.

\begin{table}[h]
  \centering
  \caption{QA1c phase anomaly on DUDEDISCO: high answer containment with zero format compliance across all OCR systems.}
  \label{tab:qa1c_anomaly}
  \begin{tabular}{l|ccc}
    \toprule
    OCR System & GT-in-Pred $\uparrow$ & ANLS $\uparrow$ & Exact Match $\uparrow$ \\
    \midrule
    Mistral OCR 3 → GPT-5 Mini & 0.579 & 0.000 & 0.000 \\
    Mistral OCR 2 → GPT-5 Mini & 0.567 & 0.004 & 0.000 \\
    Azure Intelligence → GPT-5 Mini & 0.567 & 0.000 & 0.000 \\
    \bottomrule
  \end{tabular}
\end{table}

\subsection{All raw results}

% -----------------------------
% Parsing: cosine similarity, CER, WER (by phase)
% -----------------------------
\begin{table*}[t]
  \centering
  \tiny
  \setlength{\tabcolsep}{3.5pt}
  \begin{tabular}{l|ccc|ccc|ccc}
    \toprule
    & \multicolumn{3}{c|}{P-a (OCR baseline)} & \multicolumn{3}{c|}{P-b (generic VLM extraction)} & \multicolumn{3}{c}{P-c (task-aware VLM extraction)} \\
    \textbf{Dataset} & \textbf{CER} $\downarrow$ & \textbf{WER} $\downarrow$ & \textbf{$\SCORE{CS}$} $\uparrow$ &
    \textbf{CER} $\downarrow$ & \textbf{WER} $\downarrow$ & \textbf{$\SCORE{CS}$} $\uparrow$ &
    \textbf{CER} $\downarrow$ & \textbf{WER} $\downarrow$ & \textbf{$\SCORE{CS}$} $\uparrow$ \\
    \midrule
    IAM (handwriting) & 0.0894 & 0.3011 & 0.9457 & 0.1626 & 0.2234 & 0.9046 & 0.0800 & 0.1098 & 0.9136 \\
    ICDAR (scene text) & 5.5261 & 4.6697 & 0.4547 & 2.1256 & 2.0210 & 0.5589 & 0.7336 & 0.8544 & 0.4744 \\
    RxPad (prescriptions) & 0.6541 & 0.5886 & 0.4820 & 0.6601 & 0.5935 & 0.4758 & 0.6591 & 0.5828 & 0.4818 \\
    \bottomrule
  \end{tabular}
  \caption{Parsing metrics (cosine similarity, CER, WER) across datasets and phases, extracted from the notebook summary outputs. Where multiple systems are reported within a phase, the row reflects the system with the highest cosine similarity for that phase on that dataset (and its corresponding CER/WER).}
  \label{tab:parsing_metrics_all}
\end{table*}

% -----------------------------
% QA: GT-in-Pred, ANLS, exact match (by phase)
% -----------------------------
\begin{table*}[t]
  \centering
  \tiny
  \setlength{\tabcolsep}{4pt}
  \begin{tabular}{l l l c c c}
    \toprule
    \textbf{Dataset} & \textbf{Phase} & \textbf{System} & $\SCORE{GT\text{-}in\text{-}Pred}$ $\uparrow$ & $\SCORE{ANLS}$ $\uparrow$ & $\SCORE{EM}$ $\uparrow$ \\
    \midrule
    DocVQA & $\QAcode{1}{a}$ & \texttt{azure\_intelligence\_\_gpt-5-mini} & 0.8405 & 0.6284 & 0.4622 \\
    DocVQA & $\QAcode{1}{b}$ & \texttt{azure\_intelligence\_\_gpt-5-mini} & 0.8763 & 0.8406 & 0.7444 \\
    DocVQA & $\QAcode{1}{c}$ & \texttt{azure\_intelligence\_\_gpt-5-mini} & 0.8571 & 0.7412 & 0.5714 \\
    DocVQA & $\QAcode{2}{a}$ & \texttt{gpt-5-mini\_\_gpt-5-mini} & 0.8514 & 0.6159 & 0.4438 \\
    DocVQA & $\QAcode{2}{b}$ & \texttt{gpt-5-mini\_\_gpt-5-mini} & 0.8956 & 0.8683 & 0.7791 \\
    DocVQA & $\QAcodeText{3}{fewshot}$ & \texttt{gpt-5-nano} & 0.8740 & 0.7455 & 0.4360 \\
    DocVQA & $\QAcode{3}{a}$ & \texttt{gpt-5-nano} & 0.8798 & 0.7414 & 0.5471 \\
    \midrule
    DUDE & $\QAcode{1}{a}$ & \texttt{azure\_intelligence\_\_gpt-5-mini} & 0.5173 & 0.3555 & 0.2054 \\
    DUDE & $\QAcode{1}{b}$ & \texttt{mistral\_document\_ai\_\_gpt-5-mini} & 0.4703 & 0.5120 & 0.3614 \\
    DUDE & $\QAcode{1}{c}$ & \texttt{mistral\_document\_ai\_\_gpt-5-mini} & 0.5619 & 0.0035 & 0.0000 \\
    DUDE & $\QAcode{2}{a}$ & \texttt{gpt-5-mini\_\_gpt-5-mini} & 0.4876 & 0.3451 & 0.1807 \\
    DUDE & $\QAcode{2}{b}$ & \texttt{gpt-5-mini\_\_gpt-5-mini} & 0.4431 & 0.5049 & 0.3342 \\
    DUDE & $\QAcodeText{3}{fewshot}$ & \texttt{gpt-5-mini} & 0.4888 & 0.3631 & 0.2084 \\
    DUDE & $\QAcode{3}{a}$ & \texttt{gpt-5-mini} & 0.4975 & 0.3906 & 0.2302 \\
    \midrule
    InfographicVQA & $\QA{OCR}{generic}$ & \texttt{azure\_intelligence\_\_gpt-5-mini} & 0.7347 & 0.5018 & 0.3967 \\
    InfographicVQA & $\QA{OCR}{cot}$ & \texttt{azure\_intelligence\_\_gpt-5-mini} & 0.7302 & 0.6808 & 0.6168 \\
    InfographicVQA & $\QA{OCR}{task}$ & \texttt{azure\_intelligence\_\_gpt-5-mini} & 0.7537 & 0.6290 & 0.5149 \\
    InfographicVQA & $\QA{VLM-2stage}{generic}$ & \texttt{gpt-5-nano\_\_gpt-5-nano} & 0.6366 & 0.5471 & 0.4213 \\
    InfographicVQA & $\QA{VLM-2stage}{cot}$ & \texttt{gpt-5-mini\_\_gpt-5-mini} & 0.6796 & 0.6257 & 0.5837 \\
    InfographicVQA & $\QA{VLM-2stage}{task}$ & \texttt{gpt-5-mini\_\_gpt-5-mini} & 0.7110 & 0.5846 & 0.4768 \\
    InfographicVQA & $\QA{VLM-direct}{fewshot}$ & \texttt{gpt-5-mini} & 0.7776 & 0.5252 & 0.4088 \\
    InfographicVQA & $\QA{VLM-direct}{generic}$ & \texttt{gpt-5-nano} & 0.7154 & 0.3839 & 0.2585 \\

    \bottomrule
  \end{tabular}
  \caption{QA metrics ($\SCORE{GT\text{-}in\text{-}Pred}$, $\SCORE{ANLS}$, $\SCORE{EM}$) across datasets and phases, extracted from the notebook summary outputs. For each dataset and phase, the row reflects the best run among those marked \texttt{[PRIMARY]} in the notebook outputs (selected by $\SCORE{ANLS}$, then $\SCORE{EM}$).}
  \label{tab:qa_metrics_all}
\end{table*}

\subsection{Summary}

\label{appendix:tldr}

We summarise (i) \textbf{parsing winners} using $\SCORE{CS}$ and (ii) \textbf{QA winners} using $\SCORE{GT\text{-}in\text{-}Pred}$ (the primary metrics used throughout the paper). A \emph{winner} is the approach that achieved the \textbf{best observed performance} on the corresponding dataset/metric. Entries are: \textbf{1} = wins, \textbf{0} = loses, \textbf{1--1} = comparable.

\begin{table*}[t]
  \centering
  \small
  \setlength{\tabcolsep}{8pt}

  \begin{minipage}[t]{0.48\textwidth}
    \centering
    \textbf{Parsing}\\[0.25em]
    \begin{tabular}{lcc}
      \toprule
      \textbf{Dataset} & \textbf{OCR} & \textbf{VLM} \\
      \midrule
      IAM$_{\text{DISCO}}$ & 1 & 0 \\
      ICDAR$_{\text{DISCO}}$ & 0 & 1 \\
      RxPad & 1--1 & 1--1 \\
      \midrule
      \textbf{Total} & 2 & 2 \\
      \bottomrule
    \end{tabular}
  \end{minipage}
  \hfill
  \begin{minipage}[t]{0.48\textwidth}
    \centering
    \textbf{Question answering}\\[0.25em]
    \begin{tabular}{lcc}
      \toprule
      \textbf{Dataset} & \textbf{OCR} & \textbf{VLM} \\
      \midrule
      DocVQA$_{\text{DISCO}}$ & 0 & 1 \\
      InfographicVQA$_{\text{DISCO}}$ & 0 & 1 \\
      DUDE$_{\text{DISCO}}$ & 1 & 0 \\
      \midrule
      \textbf{Total} & 1 & 2 \\
      \bottomrule
    \end{tabular}
  \end{minipage}

  \caption{Summary: which approach wins on \textbf{parsing} (primary metric: $\SCORE{CS}$) and \textbf{question answering} (primary metric: $\SCORE{GT-in-Pred}$). Totals count a comparable result (1--1) as 1 point for both approaches.}
  \label{tab:summary_wins}
\end{table*}

Table~\ref{tab:winner_summary_extended} provides a compact winner summary by dataset and evaluation criterion.

\begin{table}[h]
  \centering
  \begin{tabular}{l|ccc|c}
    \toprule
    Criterion & DocVQA & InfographicVQA & DUDE & Overall \\
    \midrule
    \multicolumn{5}{l}{\textit{Parsing Effectiveness (GT-in-Extracted-Text)}} \\
    Azure Intelligence & 1 & 1 & 1 & 3/3 \\
    Mistral OCR 2 & 0 & 0 & $\sim$ & 0/3 \\
    Mistral OCR 3 & 0 & 0 & 0 & 0/3 \\
    GPT-5 Mini (VLM) & $\sim$ & 0 & 1 & 1/3 \\
    \midrule
    \multicolumn{5}{l}{\textit{QA Performance (GT-in-Pred with best prompt)}} \\
    QA$_{\text{OCR}}$ & 0 & 0 & 1 & 1/3 \\
    QA$_{\text{VLM-2stage}}$ & 0 & 0 & 0 & 0/3 \\
    QA$_{\text{VLM-direct}}$ & 1 & 1 & 0 & 2/3 \\
    \midrule
    \multicolumn{5}{l}{\textit{Computational Efficiency (lowest latency)}} \\
    Azure Intelligence & 0 & 0 & 0 & 0/3 \\
    Mistral OCR 2 & 1 & $\sim$ & 1 & 2/3 \\
    Mistral OCR 3 & 0 & 1 & 0 & 1/3 \\
    \bottomrule
  \end{tabular}
  \caption{Winner summary by dataset and evaluation criterion. 1 = wins, 0 = loses, $\sim$ = comparable performance (within 2 percentage points).}
  \label{tab:winner_summary_extended}
\end{table}

\textbf{Conclusion.} Mistral OCR 3's consistent underperformance relative to Mistral OCR 2 across all datasets and metrics (parsing effectiveness, QA accuracy, and no latency advantage) indicates a clear regression. Practitioners deploying Mistral OCR should prefer version 2505 over 2512 until the performance issues in the newer version are resolved. More broadly, these findings underscore the importance of empirical validation: model version increments do not guarantee improved performance, and deployment decisions should be driven by benchmark results on representative data rather than version numbers or release dates.

\noindent\textbf{Conclusion.} Within the scope of our evaluation, direct VLM calls achieve the strongest performance more often overall (particularly on QA and visually structured documents), while OCR-based pipelines remain the most reliable choice on long or multi-page documents.

\clearpage
\section{Design Recommendations}
\label{appendix:recommendations}

\subsection{Document-aware pipeline selection}
Based on our findings:

\begin{enumerate}
  \item \textbf{Handwritten text:} Use specialised OCR. VLMs lag by $\sim 5$--$9\%$ even with task-aware prompting.

  \item \textbf{Multilingual documents:} Use VLMs with generic prompts. OCR systems struggle on non-Latin scripts.

  \item \textbf{Single-page visual QA:} Direct VQA finds correct answers most often. For precise formatting, consider OCR pipelines with prompt engineering.

  \item \textbf{Multi-page documents:} OCR pipelines provide more reliable grounding for complex reasoning.

  \item \textbf{Prompt design:} Start with generic prompts; task-aware prompts can degrade performance on diverse inputs.
\end{enumerate}

We highlight the metric discrepancy between $\SCORE{GT\text{-}in\text{-}Pred}$ and $\SCORE{ANLS}$ in Section~\ref{ap:limitations} and recommend using complementary metrics depending on the desired evaluation properties.
Additional figures for detailed results discussion are omitted in this anonymous version.

% (Removed stray/unfinished table fragment that was breaking compilation.)

\subsection{Time evaluation}
\label{sec:cost_time}

\begin{figure}[h]
  \centering
  \includegraphics[width=\textwidth]{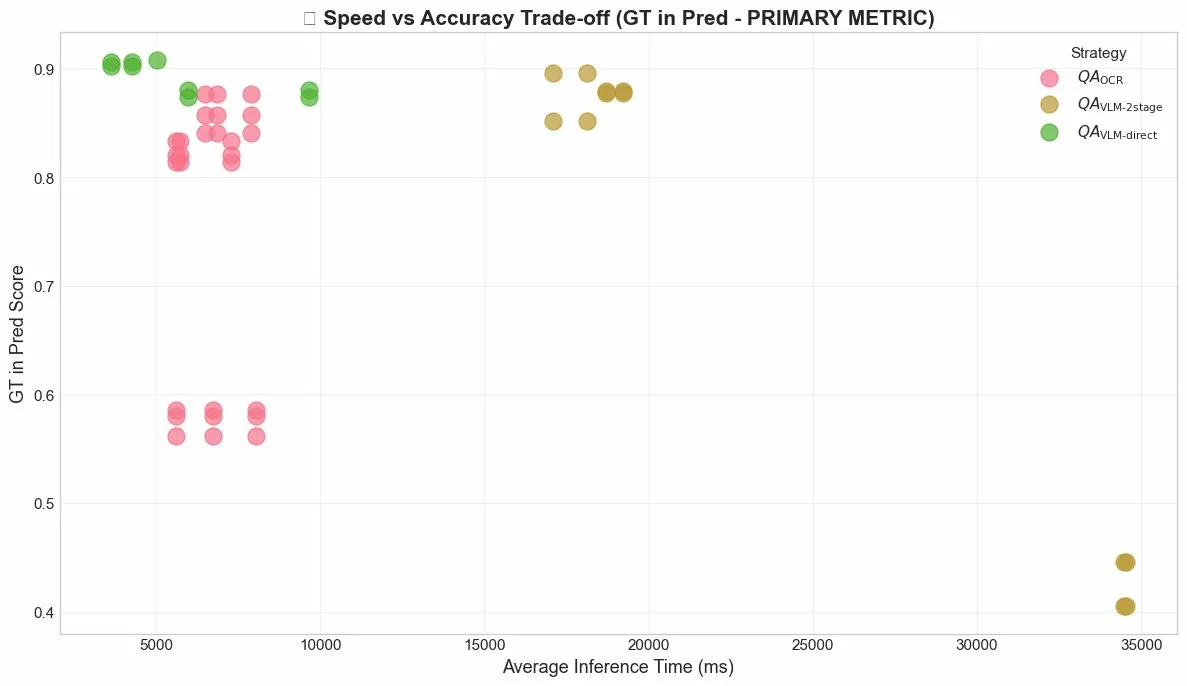}
  \caption{Speed vs accuracy trade-off across QA strategies on DocVQA$_{\text{DISCO}}$. $\QA{VLM-direct}{}$ (green) achieves the best efficiency frontier: highest accuracy with lowest latency. $\QA{VLM-2stage}{}$ (olive) incurs 2--4$\times$ longer inference times due to sequential parsing and reasoning stages.}
  \label{fig:speed_accuracy}
\end{figure}

We analyse the trade-off between inference time and accuracy across the three QA strategies on DocVQA$_{\text{DISCO}}$, InfographicVQA$_{\text{DISCO}}$, and DUDE$_{\text{DISCO}}$. As an example, figure~\ref{fig:speed_accuracy} plots average inference time (ms) against $\SCORE{GT\text{-}in\text{-}Pred}$ for InfographicVQA.

On DocVQA$_{\text{DISCO}}$, direct VQA ($\QA{VLM-direct}{}$) dominates the efficiency frontier, achieving the highest $\SCORE{GT\text{-}in\text{-}Pred}$ scores (0.87--0.91) with the fastest inference ($\sim$4--10s). OCR-based pipelines ($\QA{OCR}{}$) show comparable latency but cluster into two accuracy regimes: high performance on single-page documents (DocVQA, InfographicVQA) and lower performance on multi-page documents (DUDE). The two-stage VLM pipeline ($\QA{VLM-2stage}{}$) is consistently the slowest (17--35s), with inference time roughly doubling due to separate parsing and QA calls. When $\QA{VLM-2stage}{}$ achieves competitive accuracy, the latency cost is 2--4$\times$ that of direct VQA.

For latency-sensitive applications, direct VQA offers the best accuracy-per-millisecond ratio. Two-stage pipelines are only justified when intermediate text representations are required for downstream tasks (e.g., retrieval, audit trails) or when processing very long documents where OCR-based grounding improves reliability despite the speed penalty.

\subsection{Cost evaluation}
\label{subsec:cost}

API costs depend on pipeline architecture and number of questions per document. For a typical document ($\sim$1,500 image tokens, 500-token parsed text, 100-token answer), Table~\ref{tab:cost_comparison} compares per-document costs across strategies.

\begin{table}[h]
  \centering
  \small
  \begin{tabular}{lccc}
    \toprule
    \textbf{Strategy} & \textbf{GPT-5-nano} & \textbf{GPT-5-mini} & \textbf{Claude 3.5 Sonnet} \\
    \midrule
    $\QA{VLM-direct}{}$ (1 call) & \$0.00011 & \$0.00058 & \$0.0060 \\
    $\QA{VLM-2stage}{}$ (2 VLM calls) & \$0.00034 & \$0.00170 & \$0.0150 \\
    $\QA{OCR}{}$ (OCR + 1 LLM call)\textsuperscript{$\dagger$} & \$0.00012 & \$0.00043 & \$0.0046 \\
    \bottomrule
  \end{tabular}
  \caption{Per-document cost for single-question QA. \textsuperscript{$\dagger$}OCR cost estimated at \$0.001/page (Azure Document Intelligence); text-only LLM calls avoid image token costs.}
  \label{tab:cost_comparison}
\end{table}

For single questions, $\QA{VLM-direct}{}$ and $\QA{OCR}{}$ achieve comparable costs, while $\QA{VLM-2stage}{}$ incurs 2--3$\times$ overhead by paying for parsed text as both output and input. However, when asking \textbf{multiple questions per document}, two-stage parsing amortises the extraction cost: break-even occurs at $\sim$4--6 questions, after which $\QA{OCR}{}$ or $\QA{VLM-2stage}{}$ become more economical. For multi-question workloads, $\QA{OCR}{}$ offers the best cost-accuracy trade-off, combining cheaper text-only QA calls with the reliability advantages observed on long documents (Section~\ref{appendix:qa_analysis}).

\clearpage
\section{How we built the \textit{DISCO} benchmark suite to be representative}
\label{appendix:representative_suite}

\subsection{Dataset Creation Methodology}

\paragraph{Sampling Strategy}

We employed three primary sampling strategies depending on each dataset's characteristics:

\begin{enumerate}
  \item \textbf{Simple random sampling}: For relatively homogeneous datasets without a strong categorical structure, we applied uniform random sampling with a fixed random seed (42) to ensure reproducibility.

  \item \textbf{Stratified sampling}: For datasets with known categorical variables (e.g., question types, languages, or content types), we applied stratified sampling to preserve the proportional representation of each category.

  \item \textbf{Balanced sampling}: For datasets with extreme class imbalance or multiple languages, we enforced balanced representation by sampling equal numbers of examples from each category.
\end{enumerate}

\paragraph{Sample Size Selection}

Target sample sizes were selected based on statistical power analysis and computational constraints:

\begin{itemize}
  \item \textbf{500 samples}: We use 500 samples for most datasets. This balances reliability with computational cost. With this sample size, 95\% confidence intervals are roughly $\pm$4 percentage points, and we can detect differences of 7–9 percentage points between systems (80\% power, $\alpha=0.05$). For continuous metrics like $\SCORE{ANLS}$, we can detect effect sizes of about 0.125 standard deviations.

  \item \textbf{Smaller samples}: For very expensive or rare datasets (RxPad: 200 samples), we used the full available dataset or maximum feasible subset.

  \item \textbf{Larger samples}: For multi-faceted datasets requiring diverse coverage (VisR-Bench: 498 documents with 17,045 QA pairs), we sampled at the document level while preserving multiple questions per document.
\end{itemize}

\subsection{Text Parsing (OCR) Datasets}

\paragraph{IAM$_{\text{DISCO}}$}
\begin{itemize}
  \item \textbf{Source:} IAM Handwriting Database~\citet{marti2002iam}
  \item \textbf{Task:} Handwriting recognition (parsing)
  \item \textbf{Original size:} $\sim$11,539 images with both handwritting and handwritten together
  \item \textbf{DISCO subset size:} 500 text line samples
  \item \textbf{Sampling:} Random sampling across writers and text styles
  \item \textbf{Data format and contribution:}
    \begin{itemize}
      \item \textbf{Reference:} printed ground-truth image (\texttt{printed.png})
      \item \textbf{Input:} handwritten text line image (\texttt{handwritten.png})
      \item Pre-cropped images for consistent evaluation
    \end{itemize}

\end{itemize}

\paragraph{ICDAR$_{\text{DISCO}}$}
\begin{itemize}
  \item \textbf{Source:} ICDAR 2015 competition on robust reading~\cite{karatzas2015icdar}
  \item \textbf{Task:} Multi-lingual scene text recognition (parsing)
  \item \textbf{Original size:} $\sim$10,000 images
  \item \textbf{DISCO subset size:} 500 samples (balanced; 50 per language category)
  \item \textbf{Sampling:} Stratified balanced sampling across 10 language categories, 50 samples each
  \item \textbf{Key metadata and contribution:}
    \begin{itemize}
      \item Text transcription with language identifier
      \item Position metadata for reading order
    \end{itemize}

\end{itemize}

\paragraph{PubLayNet$_{\text{DISCO}}$}
\begin{itemize}
  \item \textbf{Source:} PubLayNet document layout dataset~\cite{zhong2019publaynet}
  \item \textbf{Task:} Document layout analysis (parsing)
  \item \textbf{Original size:} 335,703 document images
  \item \textbf{DISCO subset size:} 500 page samples
  \item \textbf{Sampling:} Random sampling from scientific publications
  \item \textbf{Layout categories:} Text, Title, List, Table, Figure

\end{itemize}

% \paragraph{VOC2007}
% \begin{itemize}
%   \item \textbf{Source:} Custom medical laboratory report dataset (Chinese)~\cite{everingham2010pascal}
%   \item \textbf{Task:} Chinese medical lab report parsing
%   \item \textbf{Original size:} 238 samples
%   \item \textbf{DISCO subset size:} 238 samples (full dataset; no subsetting)
%   \item \textbf{Sampling:} Complete dataset inclusion
%   \item \textbf{Language:} Simplified Chinese (zh-CN)
%   \item \textbf{Key metadata / annotations:}
%     \begin{itemize}
%       \item Field-level bounding boxes
%       \item Table structure annotations (\texttt{table\_no}, \texttt{cell\_row}, \texttt{cell\_line})
%       \item Cell-level text transcriptions
%       \item Spatial ordering metadata (x, y coordinates)
%     \end{itemize}

% \end{itemize}

\paragraph{RxPad} \textit{The original dataset was used for this experiment}
\begin{itemize}
  \item \textbf{Source:} French medical prescription dataset (RxPad)~\cite{everingham2010pascal}
  \item \textbf{Task:} Medical prescription parsing (French)
  \item \textbf{Original size:} 200 samples (150 training + 50 testing)
  \item \textbf{DISCO subset size:} 200 samples (full dataset; no subsetting)
  \item \textbf{Sampling:} Complete dataset inclusion from training and testing splits
  \item \textbf{Language:} French (fr)
  \item \textbf{Key annotations:} Prescriber/patient fields, medication details, dates/signatures, administrative codes
  \item \textbf{Image characteristics:} Mix of print and handwriting; structured form layouts (average resolution 1,474 × 1,995 px)

\end{itemize}

\subsection{Question Answering (VQA) Datasets}

\paragraph{DocVQA$_{\text{DISCO}}$}
\begin{itemize}
  \item \textbf{Source:} Document Visual Question Answering (DocVQA), validation split~\cite{mathew2021docvqa}
  \item \textbf{Task:} Single-page document VQA (forms, receipts, letters)
  \item \textbf{Original size:} 5,349 QA pairs
  \item \textbf{DISCO subset size:} 500 QA pairs
  \item \textbf{Sampling:} Simple random sampling with seed=42 from the validation split
  \item \textbf{Content / document types:} Scanned business documents, forms, receipts, letters, and reports (variable scan quality; occasional handwriting)
  \item \textbf{Key metadata / annotations:}
    \begin{itemize}
      \item Document ID and page number for traceability
      \item Multiple valid answer annotations (average 1.8 answers per question)
      \item Question-type metadata (layout, handwritten, figure/diagram, etc.)
      \item Document source identifiers (UCSF document collection)
    \end{itemize}

\end{itemize}

\paragraph{InfographicVQA$_{\text{DISCO}}$}
\begin{itemize}
  \item \textbf{Source:} InfographicVQA, validation split~\cite{mathew2022infographicvqa}
  \item \textbf{Task:} Single-page infographic VQA (visual--text alignment + numerical reasoning)
  \item \textbf{Original size:} 5,186 QA pairs
  \item \textbf{DISCO subset size:} 500 QA pairs
  \item \textbf{Sampling:} Simple random sampling with seed=42 from the validation split
  \item \textbf{Content types:} Infographics, data visualizations, charts, statistical graphics
  \item \textbf{Key metadata / annotations:}
    \begin{itemize}
      \item Pre-extracted OCR text from AWS Textract included in metadata
      \item Operation/reasoning type annotations (arithmetic, comparison, etc.)
      \item Longer questions on average (e.g., mean question length 14.2 words)
    \end{itemize}

\end{itemize}

\paragraph{DUDE$_{\text{DISCO}}$}
\begin{itemize}
  \item \textbf{Source:} Document Understanding Dataset and Evaluation (DUDE)~\cite{vanlandeghem2023dude}
  \item \textbf{Task:} Multi-page document QA (cross-page reasoning + localisation)
  \item \textbf{Original size:} 8,000+ QA pairs
  \item \textbf{DISCO subset size:} 404 QA pairs (kept below 500 samples for feasibility)
  \item \textbf{Sampling:} Stratified sampling across question families (with document-level capping; max 5 QAs per document) to prevent a single question type from dominating evaluation and to ensure balanced coverage of reasoning skills.
  \item \textbf{Target question-family distribution (percent \& approx. count in 404):}
    \begin{itemize}
      \item numeric\_amount: 20\% ($\sim$81)
      \item date\_time: 15\% ($\sim$61)
      \item lookup\_entity: 40\% ($\sim$162)
      \item yes\_no: 15\% ($\sim$61)
      \item multi\_hop\_other: 10\% ($\sim$40)
    \end{itemize}
  \item \textbf{Additional stratification dimensions:}
    \begin{enumerate}
      \item \textbf{Answer type}: short text, long text, numeric, boolean
      \item \textbf{Document ID}: capped to prevent over-representation
    \end{enumerate}
  \item \textbf{Content / document types:} Real-world multi-page documents (invoices, receipts, forms, letters, financial reports, scientific papers) with multilingual content

\end{itemize}

\paragraph{ChartQAPro$_{\text{DISCO}}$}
\begin{itemize}
  \item \textbf{Source:} ChartQA Professional, validation split~\cite{masry-etal-2025-chartqapro}
  \item \textbf{Task:} Chart QA (numerical + multi-step reasoning, conversational follow-ups)
  \item \textbf{Original size:} 1,948 QA pairs
  \item \textbf{DISCO subset size:} 494 QA pairs
  \item \textbf{Sampling:} Multi-dimensional stratified sampling across question type, answer type, and conversational depth
  \item \textbf{Representative distributions (preserved):}
    \begin{itemize}
      \item Question types: Factoid (55.9\%), Conversational (16.0\%), Fact Checking (12.8\%), Multiple Choice (10.7\%), Hypothetical (4.7\%)
      \item Answer types: short text (38.3\%), numeric (37.7\%), boolean (13.2\%), multiple choice (8.9\%), long text (2.0\%)
    \end{itemize}
  \item \textbf{Key metadata / annotations:}
    \begin{itemize}
      \item Multi-turn conversational samples (2--6 follow-up questions)
      \item Paragraph context present for a subset of samples (12.6\%)
      \item Temporal/year-based reasoning required for a subset of samples (4.3\%)
    \end{itemize}

\end{itemize}

\paragraph{VisR-Bench$_{\text{DISCO}}$}
\begin{itemize}
  \item \textbf{Source:} Visual Retrieval Benchmark for long-context documents (VisR-Bench)~\cite{chen2025visr-bench}
  \item \textbf{Task:} Multi-page document retrieval + question answering (IR + VQA)
  \item \textbf{Original size:} 394 documents; 17{,}045 total QA pairs ($\approx$43.2 questions per document on average)
  \item \textbf{DISCO subset size:} 498 documents (document-level sampling; question capping to 5 QAs per doc by default)
  \item \textbf{Sampling:} Document-level sampling with per-document QA capping (\texttt{qa\_per\_doc}$\leq 5$) to address the highly unbalanced original QA distribution (some documents have many more questions than others) and to balance document diversity vs. question coverage
  \item \textbf{Content types:} Figure / table / text / multilingual documents (15 languages)
  \item \textbf{Key metadata / annotations:}
    \begin{itemize}
      \item Each QA includes \texttt{page\_index} for retrieval evaluation
      \item Pre-extracted markdown available for all pages (\texttt{all\_page\_md\_str})
      \item Wide length distribution: 2--417 pages per document (mean 21.2, median 7.0)
      \item Capping is intended to (i) prevent documents with many questions from dominating evaluation, (ii) ensure fair per-document coverage, and (iii) preserve answer length distributions within each content type (figure/table/text/multilingual)
    \end{itemize}

\end{itemize}

%==============================================================================

\section{Experimental setup}

\subsection{Experiments}
\noindent\textbf{Parsing:}

\begin{itemize}
  \item \textbf{$P_{\text{OCR}}$}: OCR baseline using specialized OCR systems
  \item \textbf{$P_{\text{VLM-base}}$}: VLM baseline with generic text extraction prompts
  \item \textbf{$P_{\text{VLM-task}}$}: VLM with task-specific, domain-aware prompts
\end{itemize}
\noindent\textbf{Question Answering:}
\begin{itemize}
  \item \textbf{$QA_{\text{OCR}}$ (OCR$\to$QA)}: Specialized OCR extracts text, then VLM performs question answering
  \item \textbf{$QA_{\text{VLM-2stage}}$ (VLM$\to$QA)}: VLM extracts text from image, then same/different VLM performs question answering
  \item \textbf{$QA_{\text{VLM-direct}}$ (Direct VQA)}: Single-step end-to-end VLM directly answers question from image
\end{itemize}

\subsection{Metrics}
\label{sec:metrics}

\begin{table}[t]
  \centering
  \small
  \setlength{\tabcolsep}{8pt}
  \begin{tabularx}{\linewidth}{@{}lXX@{}}
    \toprule
    & \textbf{Parsing} & \textbf{Question answering (QA)} \\
    \midrule
    \textbf{Primary metric}
    & \textbf{$\SCORE{CS}$} (Cosine Similarity): semantic similarity between the parsed output and the ground-truth text embeddings. We use $\SCORE{CS}$ as the primary parsing metric because it captures \emph{semantic closeness} between extracted and reference text, even when the surface form differs (e.g., minor OCR noise, punctuation, or formatting changes).
    & \textbf{$\SCORE{GT-in-Pred}$} (Ground-Truth-in-Prediction): binary substring-based metric equal to 1 if the ground-truth answer string appears in the prediction, 0 otherwise. We use $\SCORE{GT-in-Pred}$ as the primary QA metric because it supports a binary correctness check and is robust to verbose LLM outputs (where the correct answer may be embedded in a longer explanation). \\
    \addlinespace
    \textbf{Other metrics}
    & \textbf{CER} (Character Error Rate): character-level edit distance normalised by ground-truth length; \ \textbf{WER} (Word Error Rate): word-level edit distance normalised by ground-truth word count.
    & \textbf{EM} (Exact Match): 1 if the prediction exactly matches the ground truth, 0 otherwise; \textbf{ANLS} (Average Normalised Levenshtein Similarity): normalised string similarity based on Levenshtein distance. \\
    \bottomrule
  \end{tabularx}
\end{table}

\subsection{Models}

\subsubsection{OCR Models}
\begin{itemize}
  \item \textbf{azure\_intelligence}:
  \item \textbf{mistral-ocr-2505}: Mistral OCR 2
  \item \textbf{mistral-ocr-2512}: Mistral OCR 3
\end{itemize}

\subsubsection{VLM Models}
\begin{itemize}
  \item \textbf{gpt-5-mini}: \texttt{gpt-5-mini} (vision-language)
  \item \textbf{gpt-5-nano}: \texttt{\texttt{gpt-5-nano}} (vision-language)
  \item \textbf{\texttt{claude-3-5-sonnet}}: \texttt{claude-3-5-sonnet} (vision-language)
\end{itemize}

\clearpage
\section{Prompts}
\label{sec:prompts}

% ==============================================================================
% PARSING TASK PROMPTS
% ==============================================================================

\subsection{Parsing Task Prompts}
\label{subsec:parsing_prompts}

\subsubsection{Phase $P_{\text{OCR}}$: OCR Baseline}
\label{prompt:p-a}

\noindent\textbf{Note}: Phase $P_{\text{OCR}}$ uses specialised OCR systems (\texttt{azure-ai-documentintelligence}, \texttt{mistral-ocr-2505}) which do not require explicit prompts. These models are trained end-to-end for text extraction and operate directly on document images.

% ------------------------------------------------------------------------------

\subsubsection{Phase $P_{\text{VLM-base}}$: VLM with Generic Prompts}
\label{prompt:p-b}

\noindent\textbf{Prompt $P_{\text{VLM-base}}$} (used for all datasets: IAM$_{\text{DISCO}}$, ICDAR$_{\text{DISCO}}$, RxPad):

\begin{quote}
  \texttt{Extract all text from this image.}
\end{quote}

% ------------------------------------------------------------------------------

\subsubsection{Phase $P_{\text{VLM-task}}$: VLM with Task-Aware Prompts}
\label{prompt:p-c}

\noindent\textbf{Prompt $P_{\text{VLM-task}}$-IAM} (handwritten documents):

\begin{quote}
  \texttt{This is a handwritten document. Extract all text carefully preserving word boundaries and maintaining the original line structure.}
\end{quote}

\vspace{0.5em}

\noindent\textbf{Prompt $P_{\text{VLM-task}}$-ICDAR} (multilingual documents):

\begin{quote}
  \texttt{This document contains text in multiple languages including Arabic, Chinese, Japanese, Korean, and Latin scripts. Extract all text from the image, preserving the original script and character encoding. Maintain spatial layout where text appears in columns or mixed directions.}
\end{quote}

\vspace{0.5em}

\noindent\textbf{Prompt $P_{\text{VLM-task}}$-VOC2007} (Chinese medical reports):

\begin{quote}
  \texttt{This is a Chinese medical laboratory report. Extract all text from the document, including all numerical values, units of measurement, and Chinese characters. Preserve the tabular structure if present.}
\end{quote}

\vspace{0.5em}

\noindent\textbf{Prompt $P_{\text{VLM-task}}$-PubLayNet} (scientific papers):

\begin{quote}
  \texttt{This is a page from a scientific paper. Extract all text from the document, preserving the section structure (title, abstract, body text, figure captions, references). Maintain paragraph breaks and list formatting.}
\end{quote}

\noindent\textit{Models tested}: \texttt{gpt-5-mini}, \texttt{gpt-5-nano}, \texttt{claude-3-5-sonnet}

% ==============================================================================
% QUESTION ANSWERING PROMPTS
% ==============================================================================

\subsection{Question Answering Task Prompts}
\label{subsec:qa_prompts}

% ------------------------------------------------------------------------------

\subsubsection{Phase $QA_{\text{OCR}}$: OCR$\to$QA Pipeline}
\label{prompt:qa1}

\noindent\textit{Stage 1}: Specialized OCR (\texttt{azure-ai-documentintelligence}, \texttt{mistral-ocr-2505} or  \texttt{mistral-ocr-2512}) extracts text without prompting.

\noindent\textit{Stage 2}: VLM answers question based on extracted text using one of three prompt variants:

\vspace{0.5em}

\noindent\textbf{Prompt $\QA{OCR}{generic}$} (generic):

\begin{quote}
  \texttt{
    Text: [extracted\_text]\\
    \\
  Answer: [question]}
\end{quote}

\vspace{0.5em}

\noindent\textbf{Prompt $\QA{OCR}{cot}$} (chain-of-thought):

\begin{quote}
  \texttt{Based on the following text, answer the question. Think step-by-step about how to find the answer.\\
    \\
    Text: [extracted\_text]\\
    \\
    Question: [question]\\
    \\
  Provide your reasoning and then the final answer.}
\end{quote}

\vspace{0.5em}

\noindent\textbf{Prompt $\QA{OCR}{task-aware}$} (task-aware + chain-of-thought):

\begin{quote}
  \texttt{You are analyzing a document. The extracted text from the document is provided below.\\
    \\
    Extracted text: [extracted\_text]\\
    \\
    Answer this specific question about the document: [question]\\
    \\
  Think step-by-step about how to find the answer. Provide your reasoning and then the final answer.}
\end{quote}

% ------------------------------------------------------------------------------

\subsubsection{Phase $QA_{\text{VLM-2stage}}$: VLM$\to$QA Pipeline}
\label{prompt:qa2}

\noindent\textit{Stage 1}: VLM extracts text from image using generic prompt (Prompt $P_{\text{VLM-base}}$, see Section~\ref{prompt:p-b}).

\noindent\textit{Stage 2}: Same or different VLM answers question using one of three prompt variants:

\vspace{0.5em}

\noindent\textbf{Prompt $\QA{VLM-2stage}{generic}$} (generic):

\begin{quote}
  \texttt{Answer: [question]}
\end{quote}

\vspace{0.5em}

\noindent\textbf{Prompt $\QA{VLM-2stage}{cot}$} (chain-of-thought):

\begin{quote}
  \texttt{Based on the text below, answer the question. Think step-by-step about how to find the answer.\\
    \\
    Text: [extracted\_text]\\
    \\
    Question: [question]\\
    \\
  Provide your reasoning and then the final answer.}
\end{quote}

\vspace{0.5em}

\noindent\textbf{Prompt $\QA{VLM-2stage}{task-aware}$} (task-aware + chain-of-thought):

\begin{quote}
  \texttt{You are analyzing a document to answer a specific question. The text extracted from the document is provided below.\\
    \\
    Document text: [extracted\_text]\\
    \\
    Question: [question]\\
    \\
    Think through the question step-by-step:
    1. Identify relevant information in the text
    2. Reason about how it answers the question
    3. Formulate your final answer\\
    \\
  Provide the final answer after your reasoning.}
\end{quote}

% ------------------------------------------------------------------------------

\subsubsection{Phase $QA_{\text{VLM-direct}}$: Direct VQA (Single-Stage)}
\label{prompt:qa3}

\noindent\textit{Single stage}: VLM directly answers question from image without explicit text extraction step. Two prompt variants:

\vspace{0.5em}

\noindent\textbf{Prompt $\QA{VLM-direct}{fewshot}$} (few-shot examples):

\begin{quote}
  \texttt{You will be shown document images and questions about them. Here are some examples:\\
    \\
    {[}Image 1{]}\\
    Question: What is the invoice total?\\
    Answer: \$1,234.56\\
    \\
    {[}Image 2{]}\\
    Question: What is the sender's name?\\
    Answer: John Smith\\
    \\
    Now answer this question about the following document:\\
    \\
    {[}Target Image{]}\\
    Question: [question]\\
  Answer:}
\end{quote}

\noindent\textit{Note}: Specific examples varied by dataset (DocVQA vs InfographicVQA) to match document types.

\vspace{0.5em}

\noindent\textbf{Prompt $\QA{VLM-direct}{generic}$} (generic):

\begin{quote}
  \texttt{Look at this document image and answer the question.\\
    \\
    Question: [question]\\
    \\
  Provide a concise answer based on the visible content in the image.}
\end{quote}

% ------------------------------------------------------------------------------

\vspace{0.5em}

% % ==============================================================================
% % SUMMARY AND DESIGN RATIONALE
% % ==============================================================================
\clearpage
\section{Qualitative examples}

\subsection{Parsing}

\subsubsection{IAM$_{\textit{DISCO}}$ (parsing)}

% Assumes \sampleblock{...} already exists.
% If not, replace \sampleblock{...} with your preferred box macro.

\noindent\textbf{\#1}\hfill
{\footnotesize $\SCORE{CER}=0.0521 \mid \SCORE{WER}=0.2381 \mid \SCORE{ANLS}=0.9491 \mid \SCORE{CS}=0.9502$}\par
\vspace{0.35em}
\textbf{Prediction:}\par
\sampleblock{He looked at her. Head thrown back in a pool of hair, her blood-red lips parted and the beating of her heart in the full throat. Her mouth did things he thought no human being could stand without dying, but he went on living in an ocean of voluptuousness, that swelled and ebbed over him, under him, in him and through him ...}

\vspace{0.9em}

\textbf{Ground truth:}\par
\sampleblock{He looked at her . Heard thrown back in a pool of hair , her blood-red lips parted and the beating of her heart in the full throat . Her mouth did things he thought no human being could stand without dying , but he went on living in an ocean of voluptuousness , that swelled and ebbed over him , under him , in him and through him ...}

\vspace{0.75em}

\noindent\textbf{\#2}\hfill
{\footnotesize $\SCORE{CER}=0.0638 \mid \SCORE{WER}=0.2344 \mid \SCORE{ANLS}=0.9371 \mid \SCORE{CS}=0.9568$}\par
\vspace{0.35em}
\textbf{Prediction:}\par
\sampleblock{Unless they do at least that, Dr. Verwoerd will be able to return home claiming a triumph. His smile will be blander than ever. WE are in for it again: another Royal Wedding. Between now and June, when the Duke of Kent will marry Miss Worsley, hardly a day will pass without a story or a picture or probably both, about the nuptial arrangements.}

\vspace{0.9em}

\textbf{Ground truth:}\par
\sampleblock{1\newline Unless they do at least that , Dr. Verwoerd will be able to return home claiming a triumph\newline His smile will be blander then ever . WE are in for it again : another Royal Wedding . Between\newline how and June , when the Duke of Kent will marry Miss Worsley, hardly a day will pass\newline without a story or a picture or probably both, about the nuptial arrangements}

\vspace{0.75em}

\noindent\textbf{\#3}\hfill
{\footnotesize $\SCORE{CER}=0.0790 \mid \SCORE{WER}=0.3279 \mid \SCORE{ANLS}=0.9265 \mid \SCORE{CS}=0.9566$}\par
\vspace{0.35em}
\textbf{Prediction:}\par
\sampleblock{``Aw, forget it'', she said cheerfully. ``I'll sting you for a double for being a naughty boy. How about the telly tomorrow afternoon?'' He felt a glow of happiness steal over him. Everything was all right now, thank God. She wasn't going to break with him, after all. For the moment it was the only thing in the world that mattered.}

\vspace{0.9em}

\textbf{Ground truth:}\par
\sampleblock{' Aw , forget it " she said cheerfully . " I'll sting\newline you for a double for being a naughty boy.\newline How about the telly tomorrow afternoon ?"\newline He felt a glow of happiness steal over himy .\newline Everything was all right now, thank God. She\newline wasn't going to break with himy , after all .\newline for the moment it was the only thing in the\newline world that mattered .}

\vspace{0.75em}

\noindent\textbf{\#4}\hfill
{\footnotesize $\SCORE{CER}=0.1300 \mid \SCORE{WER}=0.4737 \mid \SCORE{ANLS}=0.8846 \mid \SCORE{CS}=0.9153$}\par
\vspace{0.35em}
\textbf{Prediction:}\par
\sampleblock{Then the whole earth will be His Altar. ``And it shall come to pass, if 1ye shall lhearken diligently unto my commandments, which I command you this day, to love the Lord your God, and to serve Him with all your heart and with all your soul.'' This may seem very good, but there is something deficient.}

\vspace{0.9em}

\textbf{Ground truth:}\par
\sampleblock{They the whole earth will be His Altar . " And it\newline shall came to pass , if Iye shall 1hearkey\newline diligently uyto my cormaycryents, which I coupyayd\newline You this day , to love the Lord your God , and to\newline serve thing with all your heart and with all\newline your soul . "This may seem very good , but there\newline Is something deficient .}

\vspace{0.75em}

\noindent\textbf{\#5}\hfill
{\footnotesize $\SCORE{CER}=0.1000 \mid \SCORE{WER}=0.2857 \mid \SCORE{ANLS}=0.9052 \mid \SCORE{CS}=0.9585$}\par
\vspace{0.35em}
\textbf{Prediction:}\par
\sampleblock{The plain, sober manner of its style all the more tellingly points up not only the horror of the case itself, which floundered on to the electrocution four years later of a German-born Bronx carpenter named Bruno Richard Hauptmann, but to the raree-show emotionalism and sensation-hunger of that era.}

\vspace{0.9em}

\textbf{Ground truth:}\par
\sampleblock{The plain , sober manner of its\newline Style\newline all the more tellingly\newline points up not only the horror\newline of the case itself , which\newline floundered on to the electrocution\newline four years later of a German -\newline Broux carpenter named\newline bom\newline Bruno Richard Hauptmann , but\newline to the raree-show emotionalism\newline and sensation - hunger of that\newline era.}

% Assumes \sampleblock{...} exists.

\noindent\textbf{\#6}\hfill
{\footnotesize $\SCORE{CER}=0.1912 \mid \SCORE{WER}=0.3077 \mid \SCORE{ANLS}=0.8150 \mid \SCORE{CS}=0.8304$}\par
\vspace{0.35em}
\textbf{Prediction:}\par
\sampleblock{The plain, sober manner of its style all the more tellingly points up not only the horror of the case itself, which floundered on to the electrocution four years later of a German-born Bronx carpenter named Bruno Richard Hauptmann, but to the raree-show emotionalism and sensation-hunger of that era.}

\vspace{0.9em}

\textbf{Ground truth:}\par
\sampleblock{The peculiar social balance of the\newline
  style in the whole relating\newline
  points up not only the horror\newline
  of the case itself, which\newline
  flourished on to the electrocution\newline
  four years later of a german-\newline
  born bronx carpenter named\newline
  Bruno Richard Hauptmann, but\newline
  to the rase-show emotionalism\newline
  and sensation-hunger of that\newline
era.}

\vspace{0.75em}

\noindent\textbf{\#7}\hfill
{\footnotesize $\SCORE{CER}=0.1465 \mid \SCORE{WER}=0.2254 \mid \SCORE{ANLS}=0.8586 \mid \SCORE{CS}=0.8645$}\par
\vspace{0.35em}
\textbf{Prediction:}\par
\sampleblock{There is just a hope that we may uncover some weakness, and find a way of fighting back at them. Michael agreed, and suggested that they use Dan as a specimen demonstrating how the Thetans machinations had been working out. It occurred to Steve that this may not have been entirely an objective suggestion on her part; but he thought it a good idea nevertheless.}

\vspace{0.9em}

\textbf{Ground truth:}\par
\sampleblock{there is just a hope that we may uncover\newline
  some weakness and find a way of fighting\newline
  back at them. Heather agreed and suggested\newline
  that they use Dan as a specimen demonstra-\newline
  ting how the thetans manipulations had been\newline
  working out. It occurred to Steve that this may\newline
  not have been entirely an objective suggestion\newline
  on her part, but he thought it a good idea\newline
nevertheless.}

\vspace{0.75em}

\noindent\textbf{\#8}\hfill
{\footnotesize $\SCORE{CER}=0.0824 \mid \SCORE{WER}=0.0909 \mid \SCORE{ANLS}=0.9176 \mid \SCORE{CS}=0.8890$}\par
\vspace{0.35em}
\textbf{Prediction:}\par
\sampleblock{Tonight, for the first time, he had abandoned all pretence and shown her the honest desperation of his feeling for her. She had neither encouraged nor completely rejected him. In some perverse way their brief quarrel had forged a bond between them. No doubt she had every intention of keeping both of them on a string. On the whole he probably had a slight advantage over the young man, inasmuch as he had money to spend and she was a girl who had a healthy respect for the material things of life.}

\vspace{0.9em}

\textbf{Ground truth:}\par
\sampleblock{Tonight, for the first time, he had abandoned all\newline
  pretence and shown her the honest desperation of his feeling\newline
  for her. She had neither encouraged nor completely rejected\newline
  him. In some perverse way their brief quarrel had forged\newline
  a bond between them. No doubt she had every intention\newline
  of keeping both of them on a string. On the whole he probably\newline
  had a slight advantage over the young man, inasmuch as he\newline
  had money to spend and she was a girl who had a healthy\newline
respect for the material things of life.}

\vspace{0.75em}

\noindent\textbf{\#9}\hfill
{\footnotesize $\SCORE{CER}=0.0996 \mid \SCORE{WER}=0.1395 \mid \SCORE{ANLS}=0.9027 \mid \SCORE{CS}=0.8465$}\par
\vspace{0.35em}
\textbf{Prediction:}\par
\sampleblock{In Fanny the pregnant girl is befriended by an old man. Here it is a young homosexual, estranged from women but yet moved by a strong instinct that extends to the unborn child as much as to the expectant mother, who acts as a protector and comforter to her in her hour of need. He shares her room and gives her his forlorn gift of companionship and sympathy - `you need someone to love you while you are looking for someone to love'.}

\vspace{0.9em}

\textbf{Ground truth:}\par
\sampleblock{In funny the pregnant girl is befriended by an old man. Here it is a young homosexual, estranged from women but yet moved by a strong maternal instinct to the unborn child as much as to the expectant mother who acts as a protector and comforter to her in her hour of need. He shares her room and gives her his fortune gift of companionship and sympathy - ``you need someone to love you while you are looking for someone to love''.}

\vspace{0.75em}

\noindent\textbf{\#10}\hfill
{\footnotesize $\SCORE{CER}=0.0504 \mid \SCORE{WER}=0.0968 \mid \SCORE{ANLS}=0.9496 \mid \SCORE{CS}=0.9105$}\par
\vspace{0.35em}
\textbf{Prediction:}\par
\sampleblock{That is doubtful. If, however, in addition to her new good-neighbour gesture, Germany takes a really big share in giving aid to underdeveloped nations, the world outlook will be brighter. What gives rise to optimism is the sign that Germany and the other leading Western nations are at long last moving towards a solution of currency problems by co-operation.}

\vspace{0.9em}

\textbf{Ground truth:}\par
\sampleblock{That is doubtful. If however, in addition to her new good-neighbour gesture, Germany takes a really big share in giving aid to underdeveloped nations, the world outlook will be brighter. What gives rise to optimism is the sign that Germany and the other leading Western nations are at long last moving towards a solution of currency problems by co-operation.}

% --- ICDAR_DISCO samples: stacked layout with explicit blank line ---

\subsubsection{ICDAR$_{\textit{DISCO}}$ (parsing)}
Each sample reports parsing metrics, followed by the extracted text and the reference. We only report Latin-script examples in the sample gallery.\footnote{We restrict the displayed examples to Latin script to keep the PDF readable with the current font setup (especially for typewriter-styled blocks) and to avoid missing-glyph issues for non-Latin scripts.}

\noindent\textbf{\#1}\hfill
{\footnotesize $\SCORE{CER}=0.8710 \mid \SCORE{WER}=0.8529 \mid \SCORE{ANLS}=0.0000 \mid \SCORE{CS}=0.7910$}\par
\vspace{0.35em}
\textbf{Prediction:}\par
\sampleblock{MILANO\newline PALAZZO\newline MARINO\newline PIERO\newline FRANCESCA\newline DELLA\newline Misericordia\newline Madonna\newline della\newline La\newline Alessi\newline Sala\newline Palazzo\newline Marino,\newline Milano,\newline libero\newline Ingresso\newline gennaio\newline 2017\newline all'8\newline 2016\newline dicembre\newline dal\newline -\newline 6\newline www.comune.milano.it\newline 800167619\newline infoline\newline I\newline Rinascento\newline G\newline INTESA\newline CIVITA\newline PAIAZLORFALE}

\vspace{0.9em}

\textbf{Ground truth:}\par
\sampleblock{\# MILANO · PALAZZO MARINO\newline
  PIERO DELLA FRANCESCA\newline
  La Madonna della Misericordia\newline
  Milano, Palazzo Marino, Sala Alessi\newline
  dal 6 dicembre 2016 all'8 gennaio 2017 – Ingresso libero\newline
  infoline 800167619\newline
www.comune.milano.it}

% --- RxPad samples: stacked layout with explicit blank line ---
% Assumes \sampleblock{...} exists

\subsubsection{RxPad (parsing).}

\vspace{0.75em}

\noindent\textbf{\#2}\hfill
{\footnotesize $\SCORE{CER}=0.5486 \mid \SCORE{WER}=0.3719 \mid \SCORE{ANLS}=0.0000 \mid \SCORE{CS}=0.4628$}\par
\vspace{0.35em}
\textbf{Prediction:}\par
\sampleblock{location: VILLEFRANCHE SUR SAONE,\newline
  date\_of\_prescription: 11/08/2021\newline
  1 comprime matin et midi (selon besoin) -- espacer 4h min\newline
  renew: Renouveler 3 fois\newline
  product\_name: AVODART 0,5MG\newline
  product\_name: TADALAFIL 5MG\newline
  product\_name: LEVOCARNIL 100MG/ML\newline
\dots}

\vspace{0.9em}

\textbf{Ground truth:}\par
\sampleblock{VILLEFRANCHE SUR SAONE, le 11/08/2021\newline
  1 comprime matin et midi selon besoin, en espacant les prises de 4h minimum pendant 1 mois.\newline
  A Renouveler 3 fois\newline
  AVODART 0,5MG CAPS MOLLE 30\newline
\dots}

\vspace{0.75em}

\noindent\textbf{\#3}\hfill
{\footnotesize $\SCORE{CER}=0.5784 \mid \SCORE{WER}=0.3596 \mid \SCORE{ANLS}=0.0000 \mid \SCORE{CS}=0.5558$}\par
\vspace{0.35em}
\textbf{Prediction:}\par
\sampleblock{structure\_name: MAISON MEDICALE DE GARDE\newline
  VITAMINE C 1000: 1 cp/jour pendant 1 mois\newline
  MAGNE B6: 2 cp 3 fois/jour pendant 1 mois\newline
  DOLIPRANE 1000: 1 cp 3 fois/jour si douleurs ou fievre pendant 3 jours\newline
\dots}

\vspace{0.9em}

\textbf{Ground truth:}\par
\sampleblock{MAISON MEDICALE DE GARDE\newline
  VITAMINE C 1000\newline
  1 cp par jour pendant 1 mois\newline
  MAGNE B6:\newline
  2 cp 3 fois par jour pendant 1 mois\newline
\dots}

\vspace{0.75em}

\noindent\textbf{\#6}\hfill
{\footnotesize $\SCORE{CER}=0.8764 \mid \SCORE{WER}=0.8597 \mid \SCORE{ANLS}=0.0000 \mid \SCORE{CS}=0.4714$}\par
\vspace{0.35em}
\textbf{Prediction:}\par
\sampleblock{LYON, le lundi 08 avril 2019\newline
  CENTRE MEDICAL OPHTALMOLOGIQUE POINT VISION\newline
  CARTEOL 2\% LP UNIDOSES\newline
  1 goutte le matin dans les 2 yeux\newline
  OAR 1 an\newline
\dots}

\vspace{0.9em}

\textbf{Ground truth:}\par
\sampleblock{LYON, le lundi 08 avril 2019\newline
  CENTRE MEDICAL OPHTALMOLOGIQUE\newline
  CARTEOL 2\% LP UNIDOSES\newline
  1 Goutte, LE MATIN, dans les 2 yeux\newline
\dots}

\vspace{0.75em}

\noindent\textbf{\#7}\hfill
{\footnotesize $\SCORE{CER}=0.7340 \mid \SCORE{WER}=0.6792 \mid \SCORE{ANLS}=0.0000 \mid \SCORE{CS}=0.4828$}\par
\vspace{0.35em}
\textbf{Prediction:}\par
\sampleblock{
  % Dr PETIT Hardy\newline
  le 25 novembre 2022 11h25\newline
  MB-Etab 1\newline
  FLAGYL 500 mg: 1 comprime 3 fois/jour pendant 15 jours\newline
  PHOSPHALUGEL: 1 sachet matin/midi/soir\newline
  BIRODOGYL: 1 comprime matin/midi/soir pendant 7 jours\newline
\dots}

\vspace{0.9em}

\textbf{Ground truth:}\par
\sampleblock{Medecine Generale\newline
  MB-Etab 1\newline
  le 25 novembre 2022 11h25\newline
  FLAGYL 500 mg - Comprime pellicule (Voie orale)\newline
\dots}

\vspace{0.75em}

\noindent\textbf{\#8}\hfill
{\footnotesize $\SCORE{CER}=0.5860 \mid \SCORE{WER}=0.5310 \mid \SCORE{ANLS}=0.0000 \mid \SCORE{CS}=0.4417$}\par
\vspace{0.35em}
\textbf{Prediction:}\par
\sampleblock{Dr \todo{name hidden} (Pediatre)\newline
  LE 02/08/2021\newline
  Enfant 1 mois -- Poids: 3,950 Kg\newline
1) HEXYON (a 2 mois)\newline
2) PREVENAR 13 (a 2 mois)\newline
3) ROTARIX (a 2 mois)\newline
4) PARACETAMOL (si fievre apres vaccins)\newline
5) VIATOL (en cas de diarrhees)\newline
\dots}

\vspace{0.9em}

\textbf{Ground truth:}\par
\sampleblock{PEDIATRE\newline
LE 02/08/2021\newline
Enfant 1 mois\newline
Poids : 3,950 Kg\newline
1) HEXYON Susp inj ... a 2 mois\newline
\dots}

\vspace{0.75em}

\noindent\textbf{\#9}\hfill
{\footnotesize $\SCORE{CER}=0.6495 \mid \SCORE{WER}=0.5488 \mid \SCORE{ANLS}=0.0000 \mid \SCORE{CS}=0.5556$}\par
\vspace{0.35em}
\textbf{Prediction:}\par
\sampleblock{Docteur Lacramioara Vasilache (Medecine Generale)\newline
43510 CAYRES\newline
phone: 04 71 57 30 59\newline
isoptine 240 lp: 1/matin\newline
aprovel 300: 1 matin\newline
furosemide 40: 1 matin et 1/2 midi\newline
\dots}

\vspace{0.9em}

\textbf{Ground truth:}\par
\sampleblock{Medecine Generale\newline
isoptine 240 lp\ \ 1/matin\newline
aprovel 300\ \ \ \ \ 1 matin\newline
furosemide 40\ \ \ 1 matin et 1/2/midi\newline
\dots}

\vspace{0.75em}

\noindent\textbf{\#10}\hfill
{\footnotesize $\SCORE{CER}=0.5386 \mid \SCORE{WER}=0.3556 \mid \SCORE{ANLS}=0.0000 \mid \SCORE{CS}=0.6359$}\par
\vspace{0.35em}
\textbf{Prediction:}\par
\sampleblock{MEDECINE GENERALE\newline
DOLIPRANE suppositoire: 1 toutes les 6h (2 boites)\newline
DACRYOSERUM (unidoses): 1 boite\newline
RIFAMYCINE collyre: 2 gouttes matin/midi/soir 7 jours\newline
AUGMENTIN sirop: 1 dose/10 kgs matin/midi/soir 7 jours\newline
\dots}

\vspace{0.9em}

\textbf{Ground truth:}\par
\sampleblock{MEDECINE GENERALE\newline
DOLIPRANE SUPPOSITOIRE 1 TTES LES 6 H 2 BTES\newline
DACRYOSERUM 1 BTE UNIDOSES\newline
RIFAMYCINE COLLYRE 2 GOUTTES MATIN MIDI ET SOIR 7 JRS\newline
\dots}

\subsection{Question Answering}

\subsubsection{DocVQA$_{\textit{DISCO}}$ (QA)}

% --- DocVQA samples: stacked layout with explicit blank line ---
% Assumes \sampleblock{...} exists

\noindent\textbf{\#1}\hfill
{\footnotesize $\SCORE{GT-in-Pred}=1.0000 \mid \SCORE{ANLS}=0.8889 \mid \SCORE{CS}=0.8709 \mid \SCORE{EM}=0.0000 \mid \SCORE{SM}=1.0000$}\par
\vspace{0.35em}
\textbf{Prediction:}\par
\sampleblock{\$3,000.00}

\vspace{0.9em}

\textbf{Ground truth:}\par
\sampleblock{3,000.00}

\vspace{0.75em}

\noindent\textbf{\#2}\hfill
{\footnotesize $\SCORE{GT\_in\_pred}=1.0000 \mid \SCORE{ANLS}=1.0000 \mid \SCORE{CS}=1.0000 \mid \SCORE{EM}=1.0000 \mid \SCORE{SM}=1.0000$}\par
\vspace{0.35em}
\textbf{Prediction:}\par
\sampleblock{123}

\vspace{0.9em}

\textbf{Ground truth:}\par
\sampleblock{123}

\vspace{0.75em}

\noindent\textbf{\#3}\hfill
{\footnotesize $\SCORE{GT\_in\_pred}=1.0000 \mid \SCORE{ANLS}=1.0000 \mid \SCORE{CS}=1.0000 \mid \SCORE{EM}=1.0000 \mid \SCORE{SM}=1.0000$}\par
\vspace{0.35em}
\textbf{Prediction:}\par
\sampleblock{34}

\vspace{0.9em}

\textbf{Ground truth:}\par
\sampleblock{34}

\vspace{0.75em}

\noindent\textbf{\#4}\hfill
{\footnotesize $\SCORE{GT\_in\_pred}=1.0000 \mid \SCORE{ANLS}=1.0000 \mid \SCORE{CS}=1.0000 \mid \SCORE{EM}=1.0000 \mid \SCORE{SM}=1.0000$}\par
\vspace{0.35em}
\textbf{Prediction:}\par
\sampleblock{Effect of HRT or Raloxifene on Endothelial Function}

\vspace{0.9em}

\textbf{Ground truth:}\par
\sampleblock{Effect of HRT or Raloxifene on Endothelial Function}

\vspace{0.75em}

\noindent\textbf{\#5}\hfill
{\footnotesize $\SCORE{GT\_in\_pred}=1.0000 \mid \SCORE{ANLS}=1.0000 \mid \SCORE{CS}=1.0000 \mid \SCORE{EM}=1.0000 \mid \SCORE{SM}=1.0000$}\par
\vspace{0.35em}
\textbf{Prediction:}\par
\sampleblock{Godfrey Phillips India Limited}

\vspace{0.9em}

\textbf{Ground truth:}\par
\sampleblock{Godfrey Phillips India Limited}

\vspace{0.75em}

\noindent\textbf{\#6}\hfill
{\footnotesize $\SCORE{GT\_in\_pred}=1.0000 \mid \SCORE{ANLS}=0.0000 \mid \SCORE{CS}=0.6739 \mid \SCORE{EM}=0.0000 \mid \SCORE{SM}=1.0000$}\par
\vspace{0.35em}
\textbf{Prediction:}\par
\sampleblock{The memorandum is written to the ``FMI Workshop Speakers.''}

\vspace{0.9em}

\textbf{Ground truth:}\par
\sampleblock{fmi workshop speakers}

\vspace{0.75em}

\noindent\textbf{\#7}\hfill
{\footnotesize $\SCORE{GT\_in\_pred}=1.0000 \mid \SCORE{ANLS}=1.0000 \mid \SCORE{CS}=1.0000 \mid \SCORE{EM}=1.0000 \mid \SCORE{SM}=1.0000$}\par
\vspace{0.35em}
\textbf{Prediction:}\par
\sampleblock{8}

\vspace{0.9em}

\textbf{Ground truth:}\par
\sampleblock{8}

\vspace{0.75em}

\noindent\textbf{\#8}\hfill
{\footnotesize $\SCORE{GT\_in\_pred}=1.0000 \mid \SCORE{ANLS}=0.0000 \mid \SCORE{CS}=0.5310 \mid \SCORE{EM}=0.0000 \mid \SCORE{SM}=1.0000$}\par
\vspace{0.35em}
\textbf{Prediction:}\par
\sampleblock{Point 5 is ``POLYUNSATURATED FATS'' — they are recommended in \dots}

\vspace{0.9em}

\textbf{Ground truth:}\par
\sampleblock{Polyunsaturated fats}

\vspace{0.75em}

\noindent\textbf{\#9}\hfill
{\footnotesize $\SCORE{GT\_in\_pred}=1.0000 \mid \SCORE{ANLS}=1.0000 \mid \SCORE{CS}=1.0000 \mid \SCORE{EM}=1.0000 \mid \SCORE{SM}=1.0000$}\par
\vspace{0.35em}
\textbf{Prediction:}\par
\sampleblock{Robert T.\ Quittmeyer}

\vspace{0.9em}

\textbf{Ground truth:}\par
\sampleblock{Robert T.\ Quittmeyer}

\vspace{0.75em}

\noindent\textbf{\#10}\hfill
{\footnotesize $\SCORE{GT\_in\_pred}=1.0000 \mid \SCORE{ANLS}=0.0000 \mid \SCORE{CS}=0.7442 \mid \SCORE{EM}=0.0000 \mid \SCORE{SM}=1.0000$}\par
\vspace{0.35em}
\textbf{Prediction:}\par
\sampleblock{Department of Obstetrics and Gynecology (New York University\dots)}

\vspace{0.9em}

\textbf{Ground truth:}\par
\sampleblock{Department of Obstetrics and Gynecology}

\subsubsection{InfographicVQA$_{\textit{DISCO}}$ (QA)}
% --- InfographicVQA / ChartQA-style samples: stacked layout with explicit blank line ---
% Assumes \sampleblock{...} exists

\noindent\textbf{\#1}\hfill
{\footnotesize $\SCORE{GT\_in\_pred}=1.0000 \mid \SCORE{ANLS}=1.0000 \mid \SCORE{CS}=0.7026 \mid \SCORE{EM}=1.0000$}\par
\vspace{0.35em}
\textbf{Question:}\par
\sampleblock{Which is the IOT application shown in the infographic in green background color?}

\vspace{0.35em}
\textbf{Prediction:}\par
\sampleblock{WhatsApp}

\vspace{0.9em}

\textbf{Ground truth:}\par
\sampleblock{whatsapp}

\vspace{0.75em}

\noindent\textbf{\#2}\hfill
{\footnotesize $\SCORE{GT\_in\_pred}=1.0000 \mid \SCORE{ANLS}=1.0000 \mid \SCORE{CS}=1.0000 \mid \SCORE{EM}=1.0000$}\par
\vspace{0.35em}
\textbf{Question:}\par
\sampleblock{What percentage of respondents backup photos on cloud according to the survey?}

\vspace{0.35em}
\textbf{Prediction:}\par
\sampleblock{45.3\%}

\vspace{0.9em}

\textbf{Ground truth:}\par
\sampleblock{45.3\%}

\vspace{0.75em}

\noindent\textbf{\#3}\hfill
{\footnotesize $\SCORE{GT\_in\_pred}=1.0000 \mid \SCORE{ANLS}=1.0000 \mid \SCORE{CS}=1.0000 \mid \SCORE{EM}=1.0000$}\par
\vspace{0.35em}
\textbf{Question:}\par
\sampleblock{What is the ticket price (in dollars) for the quarter-finals of the 2015 ICC Cricket World Cup?}

\vspace{0.35em}
\textbf{Prediction:}\par
\sampleblock{\$50.87}

\vspace{0.9em}

\textbf{Ground truth:}\par
\sampleblock{\$50.87}

\vspace{0.75em}

\noindent\textbf{\#4}\hfill
{\footnotesize $\SCORE{GT\_in\_pred}=1.0000 \mid \SCORE{ANLS}=0.0000 \mid \SCORE{CS}=0.3686 \mid \SCORE{EM}=0.0000$}\par
\vspace{0.35em}
\textbf{Question:}\par
\sampleblock{Which topic received the second highest mentions on media?}

\vspace{0.35em}
\textbf{Prediction:}\par
\sampleblock{PPE — personal protective equipment, with 1,589 mentions.}

\vspace{0.9em}

\textbf{Ground truth:}\par
\sampleblock{ppe}

\vspace{0.75em}

\noindent\textbf{\#5}\hfill
{\footnotesize $\SCORE{GT\_in\_pred}=1.0000 \mid \SCORE{ANLS}=0.0000 \mid \SCORE{CS}=0.5260 \mid \SCORE{EM}=0.0000$}\par
\vspace{0.35em}
\textbf{Question:}\par
\sampleblock{What is total percentage of businesses laying of staff and reducing staff hours?}

\vspace{0.35em}
\textbf{Prediction:}\par
\sampleblock{68.3\% (39.9\% reduced staff hours/shifts + 28.4\% laid off staff)}

\vspace{0.9em}

\textbf{Ground truth:}\par
\sampleblock{68.3\%}

\vspace{0.75em}

\noindent\textbf{\#6}\hfill
{\footnotesize $\SCORE{GT\_in\_pred}=1.0000 \mid \SCORE{ANLS}=1.0000 \mid \SCORE{CS}=1.0000 \mid \SCORE{EM}=1.0000$}\par
\vspace{0.35em}
\textbf{Question:}\par
\sampleblock{What percentage of the survey respondents are non-travelers?}

\vspace{0.35em}
\textbf{Prediction:}\par
\sampleblock{44\%}

\vspace{0.9em}

\textbf{Ground truth:}\par
\sampleblock{44\%}

\vspace{0.75em}

\noindent\textbf{\#7}\hfill
{\footnotesize $\SCORE{GT\_in\_pred}=0.0000 \mid \SCORE{ANLS}=0.7619 \mid \SCORE{CS}=0.8440 \mid \SCORE{EM}=0.0000$}\par
\vspace{0.35em}
\textbf{Question:}\par
\sampleblock{Besides earth, which are the other two planets featured in the movie Thor?}

\vspace{0.35em}
\textbf{Prediction:}\par
\sampleblock{Asgard and Jotunheim.}

\vspace{0.9em}

\textbf{Ground truth:}\par
\sampleblock{asgard, jotunheim}

\vspace{0.75em}

\noindent\textbf{\#8}\hfill
{\footnotesize $\SCORE{GT\_in\_pred}=1.0000 \mid \SCORE{ANLS}=1.0000 \mid \SCORE{CS}=0.9335 \mid \SCORE{EM}=1.0000$}\par
\vspace{0.35em}
\textbf{Question:}\par
\sampleblock{Who is the author of \emph{Stand on Zanzibar}?}

\vspace{0.35em}
\textbf{Prediction:}\par
\sampleblock{John Brunner}

\vspace{0.9em}

\textbf{Ground truth:}\par
\sampleblock{john brunner}

\vspace{0.75em}

\noindent\textbf{\#9}\hfill
{\footnotesize $\SCORE{GT\_in\_pred}=1.0000 \mid \SCORE{ANLS}=1.0000 \mid \SCORE{CS}=0.8668 \mid \SCORE{EM}=1.0000$}\par
\vspace{0.35em}
\textbf{Question:}\par
\sampleblock{Which social media platform is ranked \#1?}

\vspace{0.35em}
\textbf{Prediction:}\par
\sampleblock{Facebook}

\vspace{0.9em}

\textbf{Ground truth:}\par
\sampleblock{facebook}

\vspace{0.75em}

\noindent\textbf{\#10}\hfill
{\footnotesize $\SCORE{GT\_in\_pred}=1.0000 \mid \SCORE{ANLS}=1.0000 \mid \SCORE{CS}=1.0000 \mid \SCORE{EM}=1.0000$}\par
\vspace{0.35em}
\textbf{Question:}\par
\sampleblock{What percentage of Americans have an e-reader as of January 2014?}

\vspace{0.35em}
\textbf{Prediction:}\par
\sampleblock{32\%}

\vspace{0.9em}

\textbf{Ground truth:}\par
\sampleblock{32\%}

\subsubsection{DUDE\texorpdfstring{$_{\textit{DISCO}}$}{_{DISCO}} (QA)}

% --- DUDE samples (QA): stacked layout with explicit blank line ---
% Replaces the table with readable, page-break-friendly blocks
% Assumes \sampleblock{...} exists

\noindent\textbf{\#1}\hfill
{\footnotesize $\SCORE{GT\text{-}in\text{-}Pred}=0 \mid \SCORE{ANLS}=0.0000$}\par
\vspace{0.35em}
\textbf{Prediction:}\par
\sampleblock{The ``ADMIN MODULES'' box on the right side --- it contains the ``Token Set Creation'' link.}

\vspace{0.9em}

\textbf{Ground truth:}\par
\sampleblock{Admin Modules.}

\vspace{0.75em}

\noindent\textbf{\#2}\hfill
{\footnotesize $\SCORE{GT\text{-}in\text{-}Pred}=1 \mid \SCORE{ANLS}=0.0000$}\par
\vspace{0.35em}
\textbf{Prediction:}\par
\sampleblock{Christ University --- 1,400 student participants.}

\vspace{0.9em}

\textbf{Ground truth:}\par
\sampleblock{CHRIST UNIVERSITY}

\vspace{0.75em}

\noindent\textbf{\#3}\hfill
{\footnotesize $\SCORE{GT\text{-}in\text{-}Pred}=0 \mid \SCORE{ANLS}=0.5333$}\par
\vspace{0.35em}
\textbf{Prediction:}\par
\sampleblock{Data for Seniors Write Wikipedia}

\vspace{0.9em}

\textbf{Ground truth:}\par
\sampleblock{Data for Seniors Write Wikipedia 2019 and first half of 2020}

\vspace{0.75em}

\noindent\textbf{\#4}\hfill
{\footnotesize $\SCORE{GT\text{-}in\text{-}Pred}=1 \mid \SCORE{ANLS}=0.0000$}\par
\vspace{0.35em}
\textbf{Prediction:}\par
\sampleblock{Down --- the vertical axis arrow points downward toward the ``Infinite Basis set'' (y-axis increases downward).}

\vspace{0.9em}

\textbf{Ground truth:}\par
\sampleblock{down}

\vspace{0.75em}

\noindent\textbf{\#5}\hfill
{\footnotesize $\SCORE{GT\text{-}in\text{-}Pred}=1 \mid \SCORE{ANLS}=0.0000$}\par
\vspace{0.35em}
\textbf{Prediction:}\par
\sampleblock{Wet --- (wet your hands).}

\vspace{0.9em}

\textbf{Ground truth:}\par
\sampleblock{Wet}

\vspace{0.75em}

\noindent\textbf{\#6}\hfill
{\footnotesize $\SCORE{GT\text{-}in\text{-}Pred}=0 \mid \SCORE{ANLS}=0.0000$}\par
\vspace{0.35em}
\textbf{Prediction:}\par
\sampleblock{A cloud.}

\vspace{0.9em}

\textbf{Ground truth:}\par
\sampleblock{anchor}

\vspace{0.75em}

\noindent\textbf{\#7}\hfill
{\footnotesize $\SCORE{GT\text{-}in\text{-}Pred}=0 \mid \SCORE{ANLS}=0.0000$}\par
\vspace{0.35em}
\textbf{Prediction:}\par
\sampleblock{Photo credit: NASA/Kim Shiflett}

\vspace{0.9em}

\textbf{Ground truth:}\par
\sampleblock{NASA/Ben Smegelsky}

\vspace{0.75em}

\noindent\textbf{\#8}\hfill
{\footnotesize $\SCORE{GT\text{-}in\text{-}Pred}=1 \mid \SCORE{ANLS}=0.0000$}\par
\vspace{0.35em}
\textbf{Prediction:}\par
\sampleblock{The file number is 000049.}

\vspace{0.9em}

\textbf{Ground truth:}\par
\sampleblock{49}

\vspace{0.75em}

\noindent\textbf{\#9}\hfill
{\footnotesize $\SCORE{GT\text{-}in\text{-}Pred}=1 \mid \SCORE{ANLS}=0.6279$}\par
\vspace{0.35em}
\textbf{Prediction:}\par
\sampleblock{WEP stands for Wikipedia Education Program.}

\vspace{0.9em}

\textbf{Ground truth:}\par
\sampleblock{WIKIPEDIA EDUCATION PROGRAM}

\vspace{0.75em}

\noindent\textbf{\#10}\hfill
{\footnotesize $\SCORE{GT\text{-}in\text{-}Pred}=1 \mid \SCORE{ANLS}=0.0000$}\par
\vspace{0.35em}
\textbf{Prediction:}\par
\sampleblock{Yes --- the document includes a Los Angeles Times link}

\vspace{0.9em}

\textbf{Ground truth:}\par
\sampleblock{Yes}

\end{document}

%% file: math_commands.tex
\usepackage{amsmath,amsfonts,bm}

\def\eqref#1{equation~\ref{#1}}
\def\1{\bm{1}}

\DeclareMathAlphabet{\mathsfit}{\encodingdefault}{\sfdefault}{m}{sl}
\SetMathAlphabet{\mathsfit}{bold}{\encodingdefault}{\sfdefault}{bx}{n}